\documentclass{article}
\usepackage{arXiv}            

\usepackage[numbers,compress,sort]{natbib}

\usepackage{duckuments}          
\usepackage{amsmath,amssymb,amsfonts,amsthm}
\usepackage{graphicx}
\usepackage{textcomp}
\usepackage{xcolor}
\usepackage{contour}

\usepackage[T1]{fontenc}    
\usepackage{url}            
\usepackage{booktabs}       
\usepackage{amsfonts}       
\usepackage{nicefrac}       
\usepackage{microtype}      
\usepackage[center]{subfigure} 
\usepackage{graphicx}  
\usepackage{amsmath}
\usepackage{xspace}
\usepackage{amssymb}
\usepackage{pifont}
\usepackage{stmaryrd} 

\usepackage{chngcntr}

\usepackage{caption}
\usepackage{enumitem}
\setitemize{noitemsep,topsep=0pt,parsep=0pt,partopsep=0pt}
\usepackage{tabularx}
\usepackage{rotating}
\usepackage{tablefootnote}
\usepackage{threeparttable}
\usepackage{xcolor}
\usepackage{wasysym}
\usepackage{multicol}
\usepackage{multirow}
\usepackage{mathtools}
\usepackage{esvect}
\usepackage{pdflscape}

\usepackage{algorithm}
\usepackage{algpseudocode}

\newtheorem{problem}{Problem}

\newtheorem{theorem}{Theorem}
\newtheorem*{proof*}{Proof}
\newtheorem{lemma}{Lemma}

\definecolor{airforceblue}{rgb}{0.36, 0.54, 0.66}
\definecolor{cobalt}{rgb}{0.0, 0.28, 0.67}
\definecolor{fgreen}{HTML}{228B22}
\newcommand{\cmark}{{\color{fgreen}\ding{51}}}%

\newcommand{\xmark}{{\color{red}\ding{55}}}%

\newcommand{\smallsection}[1]{\indent\underline{\smash{\textbf{#1:}}}}

\newcommand{\defeq}{\vcentcolon=}
\newcommand{\mat}[1]{\mathbf{#1}}
\newcommand{\matt}[1]{\mathbf{#1}^{\top}}

\newcommand{\vect}[1]{\mathbf{#1}}

\newcommand{\norm}[1]{\lVert #1 \rVert_{2}}
\newcommand{\card}[1]{\lvert #1 \rvert}
\newcommand{\set}[1]{\{#1\}}

\newcommand{\rpa}{{\scriptstyle \overset{+}{\rightarrow}}}
\newcommand{\rna}{{\scriptstyle \overset{-}{\rightarrow}}}
\newcommand{\lpa}{{\scriptstyle \overset{+}{\leftarrow}}}
\newcommand{\lna}{{\scriptstyle \overset{-}{\leftarrow}}}

\let\oldsum\sum
\renewcommand{\sum}{\displaystyle\oldsum} 

\newcommand{\ourconv}{\texttt{dsg-conv}\xspace}
\newcommand{\method}{\textsc{DINES}\xspace}
\newcommand{\methodsum}{\textsc{DINES-sum}\xspace}
\newcommand{\methodmax}{\textsc{DINES-max}\xspace}
\newcommand{\methodmean}{\textsc{DINES-mean}\xspace}
\newcommand{\methodattn}{\textsc{DINES-attn}\xspace}
\newcommand{\methodS}{\textsc{DINES-s}\xspace}
\newcommand{\methodSF}{\textsc{DINES-sp}\xspace}
\newcommand{\methodSFD}{\textsc{DINES-spd}\xspace}

\newcommand{\neigh}[0]{\mathcal{N}} 
\newcommand{\neighD}[0]{\mathcal{N}^\delta} 
\newcommand{\vertex}[0]{\mathcal{V}}
\newcommand{\edges}[0]{\mathcal{E}}
\newcommand{\loss}[0]{\mathcal{L}}

\newcommand{\lossssl}[0]{$\mathcal{L}_{\texttt{disc}}$\xspace}
\newcommand{\lambdassl}[0]{$\lambda_{\texttt{disc}}$\xspace}

\newcommand{\lambdareg}[0]{$\lambda_{\texttt{reg}}$\xspace}

\newcommand{\chn}[3]{\vect{f}_{#1, #2}^{(#3)}} 
\newcommand{\chnn}[2]{\vect{f}_{#1, #2}} 
\newcommand{\msg}[3]{\vect{m}_{#1, #2}^{#3}} 

\newcommand{\din}[0]{d_{\textnormal{\texttt{in}}}}
\newcommand{\dout}[0]{d_{\textnormal{\texttt{out}}}}

\newcommand{\innerproduct}[2]{\langle #1, #2 \rangle}
\newcommand{\ip}[2]{\innerproduct{\vect{z}_{u,#1}}{\vect{z}_{v,#2}}}

\title{Learning Disentangled Representations in Signed Directed Graphs without Social Assumptions}

\author[G. Ko et al.]{%
Geonwoo Ko\\
\institute{Jeonbuk National University}\\
\email{geonwooko@jbnu.ac.kr}\And
Jinhong Jung\thanks{Corresponding author}\footnotemark[1]\\
\institute{Jeonbuk National University}\\
\email{jinhongjung@jbnu.ac.kr}
}

\begin{document}

\maketitle

\begin{abstract}
Signed graphs are complex systems that represent trust relationships or preferences in various domains. 
Learning node representations in such graphs is crucial for many mining tasks. 
Although real-world signed relationships can be influenced by multiple latent factors, most existing methods often oversimplify the modeling of signed relationships by relying on social theories and treating them as simplistic factors. 
This limits their expressiveness and their ability to capture the diverse factors that shape these relationships.
In this paper, we propose \method, a novel method for learning disentangled node representations in signed directed graphs without social assumptions. 
We adopt a disentangled framework that separates each embedding into distinct factors, allowing for capturing multiple latent factors. 
We also explore lightweight graph convolutions that focus solely on sign and direction, without depending on social theories. Additionally, we propose a decoder that effectively classifies an edge's sign by considering correlations between the factors.
To further enhance disentanglement, we jointly train a self-supervised factor discriminator with our encoder and decoder. 
Throughout extensive experiments on real-world signed directed graphs, we show that \method effectively learns disentangled node representations, and significantly outperforms its competitors in the sign prediction task. 
\end{abstract}

\section{Introduction}
\label{sec:introduction}

A signed graph represents a network consisting of a set of nodes and a set of positive and negative edges between nodes.
Signed edges can model confrontational relations, such as trust/distrust, like/dislike, and agree/disagree; thus, signed graphs have been widely utilized to represent real-world relationships such as interpersonal interactions~\cite{Guha2004-yz,Kumar2016-lm,Kunegis2009-hw}, user-item relations~\cite{seo2022siren}, and voting patterns~\cite{leskovec2010signed,leskovec2010predicting,West2014-ij}. 
Mining signed graphs have received considerable attention from data mining and machine learning communities~\cite{tang2016survey} to develop diverse applications such as 
sign prediction~\cite{Kumar2016-lm,leskovec2010predicting}, 
link prediction~\cite{song2015efficient,xu2019link}, 
node ranking~\cite{jung2020random,li2019supervised,jung2016personalized}, 
node classification~\cite{tang2016node}, clustering~\cite{he2022sssnet}, 
anomaly detection~\cite{kumar2014accurately}, 
community mining~\cite{yang2007community,chu2016finding,tzeng2020discovering}, etc.

Node representation learning in signed graphs encodes nodes into low-dimensional vector representations or embeddings while considering the signed edges between nodes. 
These node representations serve as features that can be utilized for various tasks, including those mentioned earlier. 
For this, many researchers have poured immense effort into developing intriguing methods, which are mainly categorized into \textit{signed network embedding} and \textit{signed graph neural network} (signed GNN). 
Signed network embedding methods~\cite{xu2019link,ChenQLS18,yuan2017sne,Kim2018-aa,lee2020asine,xu2022dual} take unsupervised approaches to the problem by optimizing their own likelihood functions that aim to  place embeddings while preserving positive or negative relationships. 
On the other hand, signed GNNs~\cite{Derr2018-bs,Li2020-py,Liu2021-np,shu2021sgcl,Huang2021-wp,Jung2022-pm,Yan2023-ta,Fiorini2022-ef} extract node representations using their own graph convolutions considering edge signs, and they are jointly trained with a downstream task in an end-to-end manner. 

\begin{figure}[!t]
    \centering
    \vspace{-6mm}
    \subfigure[Balance theory]{
        \hspace{-2mm}
        \includegraphics[width=0.325\linewidth]{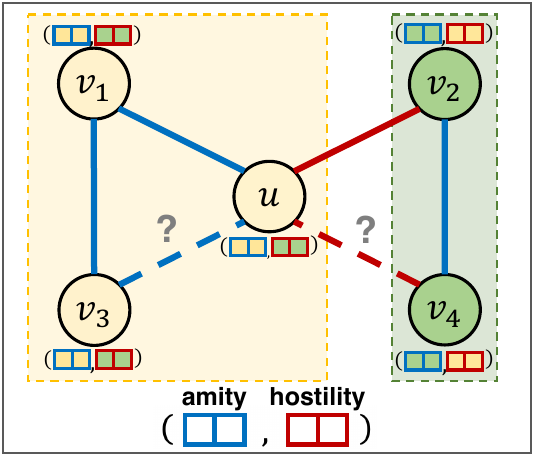}
        \label{fig:motivation:balance}
        \hspace{-3mm}
    }
    \subfigure[Status theory]{
        \includegraphics[width=0.325\linewidth]{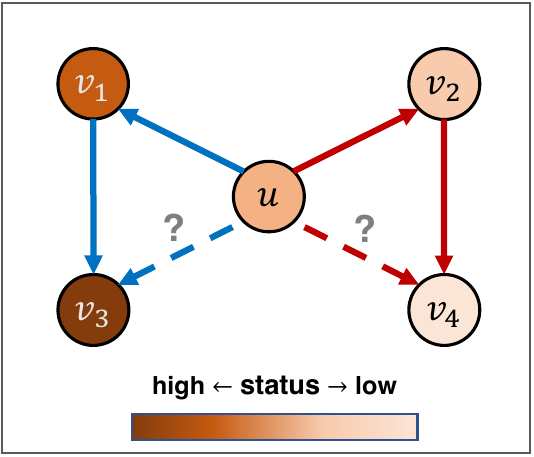}
        \label{fig:motivation:status}
        \hspace{-2mm}
    }
    \subfigure[Multiple factors]{
        \includegraphics[width=0.325\linewidth]{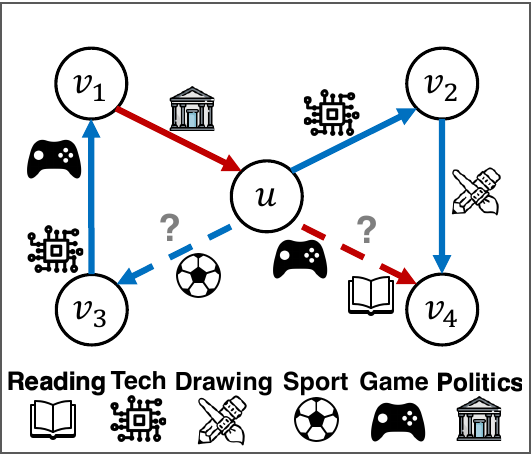}
        \label{fig:motivation:factored}
    }
    \caption{
        \label{fig:motivation}
        Illustration of how each social theory explains the sentiments of edges $(u, v_3)$ and $(u, v_4)$ where the blue edge represents a positive sentiment while the red edge represents a negative one.
        (a) According to the balance theory, $(u, v_3)$ is positive because they belong to the same group where those nodes share similar semantics for both amity and hostility factors.
        Conversely, $(u, v_4)$ is negative as they belong to different groups, resulting in opposite semantics for both factors.
        (b) The status theory explains the sentiments based on a status indicator where $v_3$ receives a positive sentiment due to its higher status compared to $u$ while $v_4$ receives a negative sentiment due to its lower status than $u$.
        (c) However, these simplistic factors derived from the social theories are insufficient to fully comprehend the sentiments, which are influenced by intricate and diverse factors.
    }
\end{figure}

However, most existing methods heavily rely on social assumptions, such as those from structural balance theory\footnote{
    The balance theory assumes that a network tends to be balanced, where individuals in the same group are more likely to share positive sentiments while those between the different groups are connected with negative sentiments, forming balanced triadic patterns such as my friend's friend is a friend or my enemy's friend is an enemy (see Figure~\ref{fig:motivation:balance}).
}~\cite{cartwright1956structural} or 
status theory\footnote{
    The status theory postulates that individuals with higher {status} tend to form positive relationships, while those with lower status may experience more negative relationships (see Figure~\ref{fig:motivation:status}).
}~\cite{leskovec2010signed}, to model node embeddings considering signed edges, which limits the expressiveness of their representations as follows:
\begin{itemize}[leftmargin=*]
	\item {
		A limitation of existing signed GNNs~\cite{Derr2018-bs,Li2020-py,Jung2022-pm} is their narrow focus on amity (or friend) and hostility (or enemy) factors, neglecting other potentially influential factors. 
		These methods only consider the information of potential friends or enemies for learning representations, adhering to the balance theory.
	}
	\item{
		Several methods~\cite{lee2020asine,Derr2018-bs,Li2020-py,Yan2023-ta} disregard the directionality of edges by treating them as undirected although the raw datasets represent directed graphs, due to the assumption of undirected graphs in the balance theory. 
		It should be noted that there is no guarantee that reciprocal edges have the same sign in real-world signed graphs~\cite{le2018characterizing}.
		
	}
	\item{
		Network embedding methods~\cite{Kim2018-aa,xu2022dual} use a set of random walk paths as input, where the signs are determined by the balance theory (i.e., $+$ sign for even negative edges; $-$ sign for odd). 
		However, the interpretation of several signs can be misleading if the balance theory is unsuitable for explaining the corresponding paths\footnote{For example, the enemy of my enemy can be uncertain as weak balance theory~\cite{davis1967clustering} states triads with all enemies are balanced.}.
	}
	\item{
		Another approach is to optimize objective functions that are specifically tailored to capture triadic relationships satisfying either the balance theory~\cite{ChenQLS18,Derr2018-bs,Li2020-py,Huang2021-wp} or the status theory~\cite{ChenQLS18,Huang2021-wp}. 
		However, this limited focus may restrict the ability of the models to capture other structural aspects of signed graphs, and secondarily, lead to inefficiency due to the computation of losses related to triadic patterns.
	}
\end{itemize}

Thus, the dependence on the social assumptions oversimplifies the relationship between nodes by considering only a few simplistic factors such as amity/hostility or status, as shown in Figures~\ref{fig:motivation:balance} and~\ref{fig:motivation:status}. 
This hinders the ability of a learning model to capture the intricate and diverse nature of real-world signed relationships influenced by multiple latent factors.
For example, Figure~\ref{fig:motivation:factored} depicts that the formation of edges in real-world signed graphs can be affected by the interaction (or correlation) between diverse factors. 
The node $u$ positively thinks of $v_3$ as they may have the same interest on \texttt{sport} factor while it negatively considers $v_4$ because their opinions disagree on \texttt{game} and \texttt{reading} factors.

Several methods~\cite{xu2019link,yuan2017sne,Liu2021-np,Fiorini2022-ef} have attempted to model node representations without relying on social theories, but their embeddings are encoded in a single embedding space, resulting in the entanglement of multiple latent factors and neglecting their potential diversity.
Disentangled GNNs~\cite{Ma2019-ce,Liu2020-yd} have emerged as a recent approach to explicitly model multiple factors in unsigned graphs, but their performance is limited when applied to signed graphs due to their lack of consideration for edge signs.
In the case of signed graphs, MUSE~\cite{Yan2023-ta} has focused on modeling multi-facet embeddings. However, its performance improvement comes from attention that considers high-order potential friends and enemies based on the balance theory, limiting scalability due to the significant increase in the number of multi-order neighbors as orders (or hops) increased.

\textit{How can we effectively learn multi-factored representations in a signed directed graph without any social assumptions?}
To address this question, we propose \method (\underline{Di}sentangled \underline{Ne}ural Networks for \underline{S}igned Directed Graphs), a new GNN method for learning node representations in signed directed graphs without the social theories. 
To model multiple latent factors, we adopt a disentangled framework in our encoder design, which separates the embedding into distinct and disentangled factors. 
We then explore lightweight and simple graph convolutions that aggregate information from neighboring nodes only considering their directions and signs.
On top of that, we propose a novel decoding strategy that constructs the edge feature by considering pairwise correlations among the disentangled factors of the nodes involved in the edge. 
We further enhance the disentanglement of factors by jointly optimizing a self-supervised factor discriminator with our encoder and decoder.
Importantly, our approach explicitly models multiple latent factors without reliance on any social assumptions, thereby avoiding the aforementioned limitations, including the degradation of expressiveness and computational inefficiency associated with the use of the social theories.

We show the strengths of \method through its complexity analysis and extensive experiments conducted on 5 real-world signed graphs with link sign prediction task, which are summarized as follows:
\begin{itemize}[leftmargin=*]
	\item {\textbf{Accurate:} 
	It achieves higher performance than the most accurate competitor, with \textbf{improvements of up to 3.1\% and 6.5\%} in terms of AUC and Macro-F1, respectively, in predicting link signs.
	}
	\item {\textbf{Scalable:} 
	Its training and inference time \textbf{scales linearly} with the number of edges in real-world and synthetic signed graphs.
	}
	\item {\textbf{Speed-accuracy Trade-off:} 
	It exhibits \textbf{a better trade-off} between training time and accuracy compared to existing signed GNN methods. 
	}
\end{itemize}

\begin{table}[!t]
	\centering
	\small
	\caption{\centering Symbols.}
	\label{tab:symbols}
	\begin{tabular}{cl}
	    \hline
		\toprule
		\textbf{Symbol} & \textbf{Definition} \\
        \midrule
        $\mathcal{G}=(\mathcal{V}, \mathcal{E})$ & {\small signed directed graph (or signed digraph)} \\
         & {\small where $\mathcal{V}$ and $\mathcal{E}$ are sets of nodes and signed edges, resp.} \\
        $n = \card{\mathcal{V}}$ & {\small number of nodes} \\
        $m = \card{\mathcal{E}}$ & {\small number of edges} \\
        $K$ & {\small number of factors} \\
        $L$ & {\small number of layers} \\
        $\din, \dout, d_{l}$ & {\small dimensions of input, output, and $l$-th embedding, respectively}\\
        $\vect{x}_{u}$ & {\small initial features of node $u$ where $\vect{x}_{u} \in \mathbb{R}^{\din}$} \\
        $\sigma(\cdot)$ & {\small non-linear activation function (e.g., \texttt{tanh})} \\
        $\mathcal{D}$ & {\small set of signed directions (or types of neighbor)} \\
        $\neigh_{u}^{\delta}$ & {\small set of neighbors of node $u$ for a given type $\delta \in \mathcal{D}$} \\
        $\vect{z}_{u}$ & {\small disentangled representation of node $u$, i.e., $\vect{z}_{u} = (\vect{z}_{u, 1}, \cdots, \vect{z}_{u,K})$}\\
        & {where $\vect{z}_{u, k} \in \mathbb{R}^{\dout/K}$ is $k$-th disentangled factor}\\
        $\mat{H}_{uv}$ & {\small $K \times K$ matrix representing the edge feature map of $u \rightarrow v$}\\
        $p_{uv}$ & {\small predicted probability of the sign of $u \rightarrow v$} \\
        \lambdassl, \lambdareg & {\small strengths of discriminative and regularization losses, resp.} \\
        $\eta$ & {\small learning rate of an optimizer} \\
		\bottomrule
		\hline
	\end{tabular}
    \vspace{-3.5mm}
\end{table}

\def\UrlFont{\bfseries\ttfamily}
For \textbf{reproducibility}, the code and the datasets are publicly available
at \url{https://github.com/geonwooko/DINES}.
The rest of the paper is organized as follows.
We provide related work and preliminaries in Sections~\ref{sec:related}~and~\ref{sec:prelim}, respectively.
We present \method in Section~\ref{sec:proposed}, followed by our experimental results in Section~\ref{sec:experiments}.
Finally, we conclude in Section~\ref{sec:conclusion}. The symbols used in this paper are summarized in Table~\ref{tab:symbols}.

\section{Related Work}
\label{sec:related}
We review related work on representation learning for signed graphs, including approaches with and without social assumptions, as well as work on disentangled GNN for unsigned graphs.
We further compare existing methods and \method w.r.t. various properties in Appendix~\ref{appendix:properties}.

\smallsection{Learning with social assumptions}
Several learning models utilize the balance theory in constructing their inputs. 
SIDE~\cite{Kim2018-aa} assigns a positive or negative sign to each node pair in truncated random walks, depending on whether the number of negative edges in the path is even or odd, respectively.
Similarly, DDRE~\cite{xu2022dual} employs random walk paths based on the balance theory as input for their expected matrix factorization with cross-noise sampling.
For adversarial learning, ASiNE~\cite{lee2020asine} generates fake edges based on the balance theory. 
SGCL~\cite{shu2021sgcl} adopts contrastive learning based on graph augmentations using the balance theory.


In signed GNNs, most previous methods have relied on the balance theory to integrate edge signs into graph convolutions.
SGCN~\cite{Derr2018-bs}, an extension of unsigned GCN~\cite{KipfW17}, leverages the balance theory to model two representations, one for potential friends and another for potential enemies.
SNEA~\cite{Li2020-py} extends SGCN by adopting an attention mechanism to capture more important neighbors.
SIDNET~\cite{Jung2022-pm} addresses the over-smoothing issue of GCN~\cite{gasteiger2018predict} by using signed random walks~\cite{jung2016personalized} that adhere to the balance theory.
MUSE~\cite{Yan2023-ta} exploited a multi-facet attention mechanism on multi-order potential friends and enemies suggested by the balance theory. 

Moreover, the social theories have been utilized in formulating structural loss functions.
BESIDE~\cite{ChenQLS18} is a deep neural network that jointly optimizes the likelihood functions of being triangles and bridge edges (not forming triangles) based on the balance and status theories. 
SDGNN~\cite{Huang2021-wp} is a signed GNN that also  optimizes its objective functions of triangles based on the social theories. 
SGCN and SNEA additionally optimize a loss function based on extended structural balance
theory~\cite{Wang2017-jw}.

As discussed in Section~\ref{sec:introduction}, these methods depend on the social theories for constructing inputs, designing graph convolutions, and formulating loss functions, which limits their expressiveness and ability to capture diverse latent factors inherent in signed relationships.

\smallsection{Learning without social assumptions}
There have been a few methods proposed to learn node representations without the balance and status theories. 
SNE~\cite{yuan2017sne} adopts a log-bilinear model considering sign-typed embeddings, and optimizes the likelihood over truncated random walks.
SLF~\cite{xu2019link} extends a latent factor model by considering sign and direction. 
Liu et al.~\cite{Liu2021-np} argued that the balance theory is too ideal to represent real-world social networks as it assumes a fully balanced graph is divided into only two groups. 
To address this, they introduced the $k$-group theory, allowing a graph to have up to $k$ groups, and proposed GS-GNN that learns node embeddings based on the $k$-group theory. 
SigMaNet~\cite{Fiorini2022-ef} introduces the Sign-Magnetic Laplacian to satisfy desirable properties of spectral GCNs for encoding both sign and direction (refer to~\cite{Fiorini2022-ef} for details).

Although these methods learn node embeddings without using the balance and status theories, they may not effectively capture the diverse latent factors that influence real-world signed relationships. This is because they bundle information from a node's neighborhood into a single holistic embedding, resulting in the entanglement of the latent factors and disregarding their potential diversity.

\smallsection{Disentangled graph neural networks}
Several methods have been recently devised for modeling disentangled representations in unsigned graphs. 
DisenGCN~\cite{Ma2019-ce} models multi-factored embeddings for each node, and performs neighborhood routing to disentangle node representations.
DGCF~\cite{Wang2020-jf} extends the disentangled approach to graph collaborative filtering so that it captures latent intents of relationships between users and items. 
DisGNN~\cite{zhao2022exploring} employs such a strategy to model disentangled edge distributions.
In addition to those methods, there are a few techniques proposed to encourage disentanglement based on independence~\cite{Liu2020-yd} or self-supervision~\cite{zhao2022exploring}.

However, such GNNs are designed for only unsigned graphs, i.e., their graph convolutions do not consider edge signs, and thus their resulting representations have limited capacity for learning signed graphs.

\begin{figure}[!t]
    \centering
    \subfigure{
    	\includegraphics[width=0.4\textwidth]{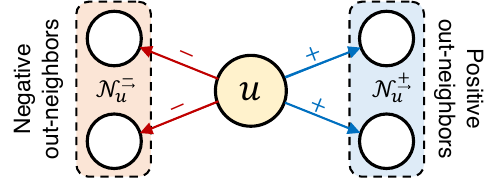}
    	\label{fig:sdn:out}
    }
    \subfigure{
    	\includegraphics[width=0.4\textwidth]{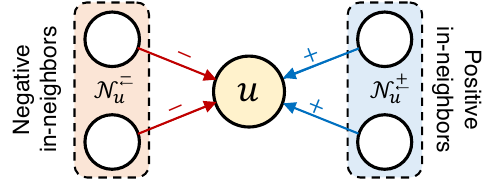}
    	\label{fig:sdn:in}
    }
    \caption{
        Concept of signed directed neighbors of node $u$.
    }
    \label{fig:sdn}
\end{figure}

\section{Preliminaries}
\label{sec:prelim}
In this section, we introduce basic notations and a formal definition of the considered problem.

\subsection{Basic Notations}
\label{sec:prelim:notations}
\smallsection{Signed directed graph}
A signed directed graph $\mathcal{G} = (\mathcal{V}, \mathcal{E})$ consists of a set $\mathcal{V}$ of nodes and a set $\mathcal{E}$ of signed directed edges (i.e., each edge has a direction and a sign of positive or negative).
We let $n=|\mathcal{V}|$ and $m=|\mathcal{E}|$ be the numbers of nodes and edges, respectively.

\smallsection{Signed directed neighbors}
In a signed directed graph, the neighbors of a given node $u$ are categorized by signs ($+$ and $-$) and directions ($\leftarrow_{(\textnormal{\texttt{in}})}$ and $\rightarrow_{(\textnormal{\texttt{out}})}$)~\cite{Huang2021-wp}.
Suppose $\mathcal{D} = \set{\rpa, \rna, \lpa, \lna}$ is the set of signed directions (e.g., $\rpa$ is the out-going direction with $+$ sign). 
Then, $\neigh_{u}^{\delta}$ denotes the set of signed directed neighbors of node $u$ with type $\delta \in \mathcal{D}$, and a node in $\neigh_{u}^{\delta}$ is called $\delta$-neighbor of node $u$. 
The set of all neighbors of node $u$, regardless of directions and signs, is denoted by $\neigh_{u} = \bigcup_{\delta \in \mathcal{D}}\neigh_{u}^{\delta}$.

We illustrate the example of signed directed neighbors of node $u$ in Figure~\ref{fig:sdn}. 
The sets of positive out-neighbors and in-neighbors are denoted by $\neigh_{u}^{\rpa}$ and $\neigh_{u}^{\lpa}$, respectively. 
The negative counterparts are expressed as $\neigh_{u}^{\rna}$ and $\neigh_{u}^{\lna}$, respectively.

\subsection{Problem Definition}
\label{sec:prelim:problem}

We describe the formal definition of the problem addressed in this paper as follows:
\begin{problem}
\label{prob:target}
\textnormal{\textbf{(Disentangled Node Representation in Signed Directed Graphs)}}
Given a signed digraph $\mathcal{G}$ and an initial feature vector $\vect{x}_{u} \in \mathbb{R}^{\din}$ for each node $u$, the problem is to learn a disentangled node representation $\vect{z}_{u}$, expressed as a list of disentangled factors, i.e., $\vect{z}_{u} = (\vect{z}_{u, 1}, \cdots, \vect{z}_{u,K})$ such that $\vect{z}_{u, i}\neq\vect{z}_{u, j}$ for $i \neq j$
where $\vect{z}_{u,i}$ is the $i$-th factor of size $\frac{\dout}{K}$, 
$d_{*}$ is the dimension of embeddings, and $K$ is the number of factors.
\end{problem}

\section{Proposed Method}
\label{sec:proposed}

\begin{figure*}[!t]
    \centering
    \includegraphics[width=\textwidth]{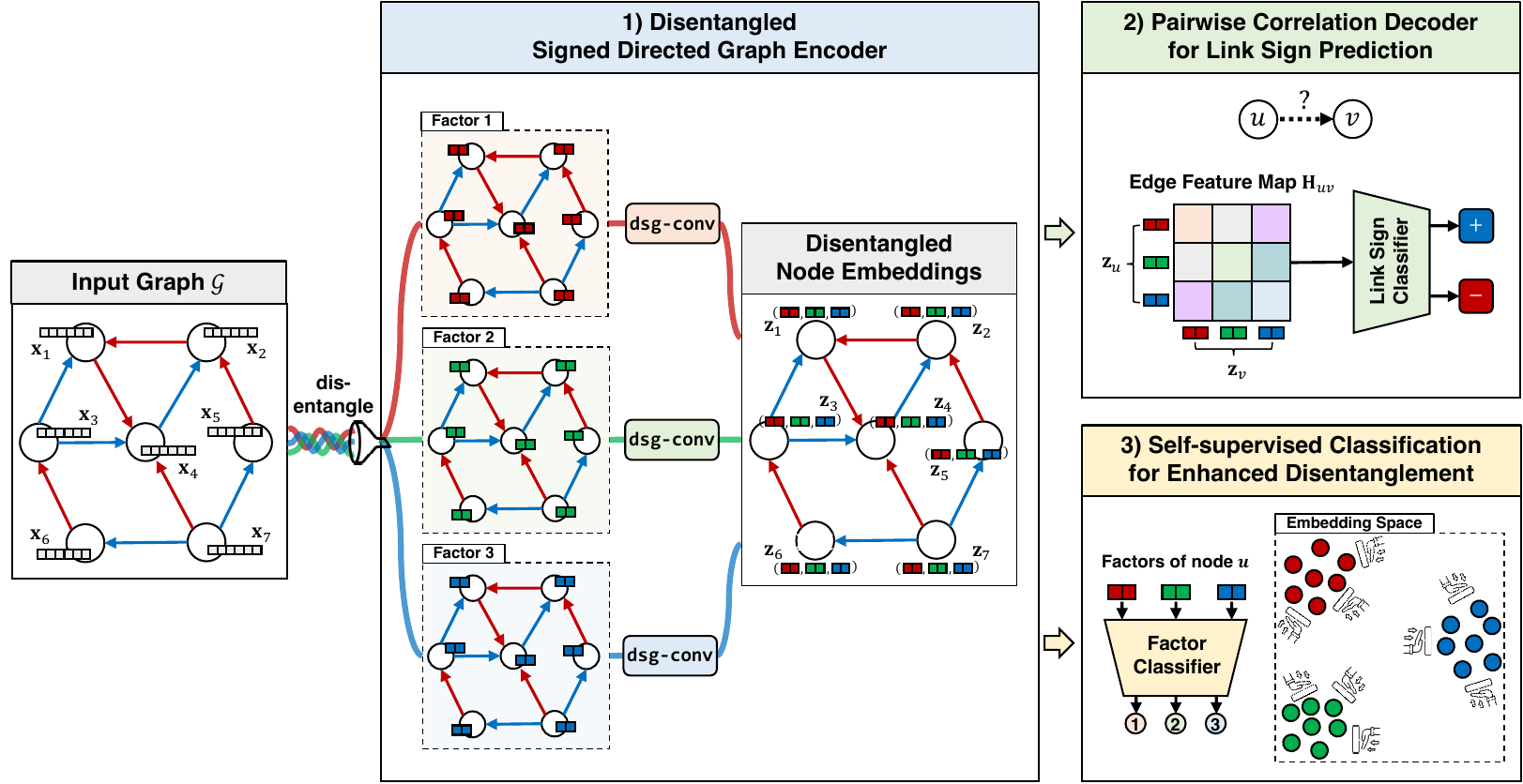}
    \caption{
        Overall structure of our proposed \method which consists of 1) disentangled signed directed graph encoder, 2) pairwise correlation decoder, and 3) self-supervised classification where the number $K$ of factors is $3$.
    }
    \label{fig:overview}
\end{figure*}

Figure~\ref{fig:overview} depicts the overall architecture of \method which consists of three modules: 1) disentangled signed directed graph encoder (Section~\ref{sec:proposed:encoder}), 2) pairwise correlation decoder (Section~\ref{sec:proposed:decoder}), and 3) self-supervised factor-wise classification (Section~\ref{sec:proposed:ssl}).

Given a signed directed graph $\mathcal{G}$ and an initial feature vector $\vect{x}_u$ for each node $u$, our encoder learns the disentangled representation (or embedding) $\vect{z}_{u}$. 
Our encoder first disentangles $\vect{x}_u$ into $K$ latent factors, and then performs disentangled signed graph convolution (\ourconv) layers for each factor space so that our model considers diverse latent factors for learning complicated relationships between nodes.
The \ourconv layer considers only directions and signs, i.e., it does not exploit any of the social theories for learning $\vect{z}_{u}$.

The disentangled embeddings are then fed forward to our decoder that predicts the sign of a given edge $u \rightarrow v$.
To capture the influence of latent factors on the relationship between $u$ and $v$, our decoder builds $\mat{H}_{uv} \in \mathbb{R}^{K \times K}$ called \textit{edge feature map} which is based on the pairwise correlations between $\vect{z}_{u}$ and $\vect{z}_{v}$.
We further adopt a self-supervised approach to enhance the disentanglement of factors by classifying each factor of a node.
Note that all of the modules are jointly trained in an end-to-end manner to effectively learn the node representations and downstream tasks.

\subsection{Disentangled Signed Directed Graph Encoder}
\label{sec:proposed:encoder}

We describe our disentangled signed directed graph encoder. 
To explicitly model multiple latent factors, we adopt the disentangled learning framework in DisenGCN~\cite{Ma2019-ce}, originally designed for unsigned graphs. 
We extend this framework by incorporating simple and lightweight signed graph convolutions, allowing us to effectively encode distinct factors without relying on any social assumptions.
Given a signed digraph $\mathcal{G}$ and the initial feature $\vect{x}_{u} \in \mathbb{R}^{\din}$ for each node $u$, our encoder aims to learn a disentangled representation $\vect{z}_{u} = (\vect{z}_{u,1}, \cdots, \vect{z}_{u,K})$, as shown in Figure~\ref{fig:overview}, where $\vect{z}_{u,i} \in \mathbb{R}^{\frac{\dout}{K}}$ is the $i$-th disentangled factor, and $K$ is the number of factors.
The overall procedure of the encoder is described in Algorithm~\ref{alg:encoder}.

\begin{figure}[!t]
\begin{algorithm}[H]
    \small
    \caption{Encoder of \method for node $u$}
    \label{alg:encoder}
    \begin{algorithmic}[1]
        \Require 
        \Statex Set of signed directions: $\mathcal{D} = \set{\rpa, \rna, \lpa, \lna}$
        \Statex Feature vector of node $u$: $\vect{x}_u \in \mathbb{R}^{d_{in}}$
        \Statex Feature vectors of neighbors of $u$: $\set{\vect{x}_v : v \in \neigh_{u}}$ where $\neigh_{u} = \bigcup_{\delta \in \mathcal{D}}\neigh_{u}^{\delta}$ 
        \Statex Number of factors: $K$ 
        \Statex Number of layers: $L$
        \Ensure 
        \Statex Disentangled representation of node $u$: $\vect{z}_{u} = (\vect{z}_{u, 1}, \cdots, \vect{z}_{u,K})$ where $\vect{z}_{u,k} \in \mathbb{R}^{\frac{d_{L}}{K}}$
        \For{\textbf{each} $v \in \neigh_{u} \cup \set{u} $} \label{alg:encoder:initialize:start}
            \For{$k$ $=$ $1$ $to$ $K$} {\color{cobalt}\Comment{\textbf{\footnotesize separate $\vect{x}_{v}$ into $K$ factors}}}
                \State $\chn{v}{k}{0} \leftarrow \texttt{normalize}\Big(\sigma\big(\texttt{FC}_{k}^{(0)}(\vect{x}_{v})\big)\Big)$
                {\color{cobalt}\Comment{\textbf{\footnotesize Equation~\eqref{eq:initial_dis}}}}
            \EndFor
        \EndFor \label{alg:encoder:initialize:end}
        \For{$k$ $=$ $1$ $to$ $K$} {\color{cobalt}\Comment{\textbf{\footnotesize aggregate neighbors for each factor through $L$ layers}}}
        	\label{alg:encoder:agg:start} 
            \For{$l$ $=$ $1$ $to$ $L$}
                    \State $\chn{u}{k}{l} \leftarrow \texttt{dsg-conv}_{k}^{(l)}\big(\big\{\chn{v}{k}{l-1} : v \in \mathcal{N}_{u} \cup \set{u} \big\}\big)$ \label{alg:encoder:dsgconv}
            \EndFor
        \EndFor \label{alg:encoder:agg:end}
        \State $\vect{z}_{u, k} \leftarrow \chn{u}{k}{L}$ \textbf{for} $1 \leq k \leq K$ 
        {\color{cobalt}\Comment{\textbf{\footnotesize $\vect{z}_{u, k}$ is a column vector of $\frac{d_{L}}{K}$ size}}} \label{alg:encoder:set}
        \State \textbf{return} $\vect{z}_{u} = (\vect{z}_{u, 1}, \cdots, \vect{z}_{u,K})$ \label{alg:encoder:return}
        {\color{cobalt}\Comment{\textbf{\footnotesize list of disentangled factors of node $u$}}}
        \item[]
        \Function{\textnormal{\texttt{dsg-conv}}$_{k}^{(l)} \big( \big\{\chn{v}{k}{l-1} : v \in \mathcal{N}_{u} \cup \set{u} \big\} \big)$}{}\label{alg:encoder:dsgconv:start}
        	\State $\msg{u}{k}{(l)} \leftarrow \bigparallel_{\delta \in \mathcal{D}}\texttt{aggregate}_k\big(\big\{\chn{v}{k}{l-1} :v \in \neighD_u\big\}\big)$\label{alg:encoder:dsgconv:message}
            \State $\chn{u}{k}{l} \leftarrow \texttt{update}_k\big(\chn{u}{k}{l-1}, \msg{u}{k}{(l)}\big)$ \label{alg:encoder:dsgconv:update}
            \State \textbf{return} $\chn{u}{k}{l} \leftarrow \texttt{normalize}(\chn{u}{k}{l})$ \label{alg:encoder:dsgconv:return}
        \EndFunction\label{alg:encoder:dsgconv:end}
    \end{algorithmic}
\end{algorithm}
\end{figure}

\smallsection{Initial disentanglement (lines~\ref{alg:encoder:initialize:start}-\ref{alg:encoder:initialize:end})} 
To model various latent factors between nodes, we first separate the initial node features into $K$ different factors. 
Specifically, for each node $v \in \mathcal{N}_{u} \cup \set{u}$, our encoder uses \texttt{FC} (fully-connected) layers to transform the feature vector $\vect{x}_{v}$ into distinct subspaces as follows:
\begin{align}
	\label{eq:initial_dis}
	\chn{v}{k}{0} = \texttt{normalize}\Big(\sigma\big(\texttt{FC}_{k}^{(0)}(\vect{x}_{v})\big)\Big) = \frac{\texttt{tanh}(\matt{W}_k \vect{x}_v + \vect{b}_k)}{\lVert \texttt{tanh}(\matt{W}_k \vect{x}_v + \vect{b}_k)\rVert_{2}}
\end{align}
where 
$\chn{v}{k}{l}$ is the $k$-th latent factor of node $v$ at step $l$, 
$\sigma(\cdot)$ is a non-linear activation function, 
$\mat{W}_{k} \in \mathbb{R}^{\din \times \frac{d_{0}}{K}}$ is the weight parameters, and $\vect{b}_{k} \in \mathbb{R}^{\frac{d_{0}}{K}}$ is the bias parameters in the aspect of the $k$-th factor.
Note that for all $k$, these parameters are differently set by random initialization before training, and the factors receive different supervision feedback from a decoder; thus, each subspace leads to distinct factors.  

For $\sigma(\cdot)$ and $\texttt{normalize}(\cdot)$, any activation and normalization functions can be used, but we use the hyperbolic tangent $\texttt{tanh}(\cdot)$ and the $\ell_2$-normalization due to the following reasons:
\begin{itemize}[leftmargin=*]
    \item{
        The $\texttt{tanh}(\cdot)$ allows each embedding to have values in a wider range compared to other functions such as $\texttt{ReLU}$ and $\texttt{sigmoid}$, and it aids in increasing the difference in predicted scores between positive and negative edges.
        Thus, it has been commonly used in previous signed GNNs~\cite{Derr2018-bs,Li2020-py,shu2021sgcl,Huang2021-wp, Jung2022-pm} to enhance the performance of link sign prediction. 
    }
    \item{
        The $\ell_2$-normalization can prevent each embedding from exploding or collapsing, while also avoiding the risk that neighboring nodes with rich features overwhelm predictions, thereby ensuring numerical stability during training.
        Because of this, it has been widely used in existing disentangled GNNs~\cite{Ma2019-ce, Liu2020-yd}.
    }
\end{itemize}

\smallsection{Disentangled signed graph convolution (lines~\ref{alg:encoder:dsgconv:start}-\ref{alg:encoder:dsgconv:end})}
The initial factor $\chn{v}{k}{0}$ of node $v$ is inadequate as the final disentangled factor $\vect{z}_{v,k}$ since it solely relies on the raw feature $\vect{x}_{v}$, and its information can be incomplete in the real world.
Thus, we need to gather information from the neighboring nodes of $v$ so that the final factor comprehensively captures the underlying semantics of the node. 

For this purpose, we design a disentangled signed graph convolution which is represented as a \ourconv layer 
 (lines~\ref{alg:encoder:dsgconv:start}-\ref{alg:encoder:dsgconv:end}):
 \begin{align}
    \label{eq:signedconv}
    \vect{f}_{u,k}^{(l)} = \ourconv_{k}^{(l)}\big(\big\{\vect{f}_{v,k}^{(l-1)} : v \in \mathcal{N}_{u} \cup \set{u} \big\} \big)
\end{align}
where $l$ indicates the $l$-th layer, and $\chn{v}{k}{l-1}$ is the $k$-th factor of neighbor $v$ of node $u$ from layer $l-1$. 

The convolution layer consists of 1) neighborhood aggregation, 2) update, and 3) normalization procedures, which are generally represented as follows:
\begin{align}
    \msg{u}{k}{(l)} &= \bigparallel_{\delta \in \mathcal{D}}
    \texttt{aggregate}_{k}\big(\big\{\chn{v}{k}{l-1} :v \in \neighD_u\big\}
    \big) \label{eq:dsg:aggregate} \\
    \chn{u}{k}{l} &= \texttt{update}_k\big(\chn{u}{k}{l-1}, \msg{u}{k}{(l)}\big) =
\sigma\Big(\texttt{FC}_{k}^{(l)}\big([\chn{u}{k}{l-1} \parallel \msg{u}{k}{(l)}]\big)\Big) \\
    \chn{u}{k}{l} &= \texttt{normalize}(\chn{u}{k}{l}) = \chn{u}{k}{l}/\norm{\chn{u}{k}{l}}
\end{align}

\begin{figure}[!t]
    \centering
    \subfigure{
    	\includegraphics[width=\textwidth]{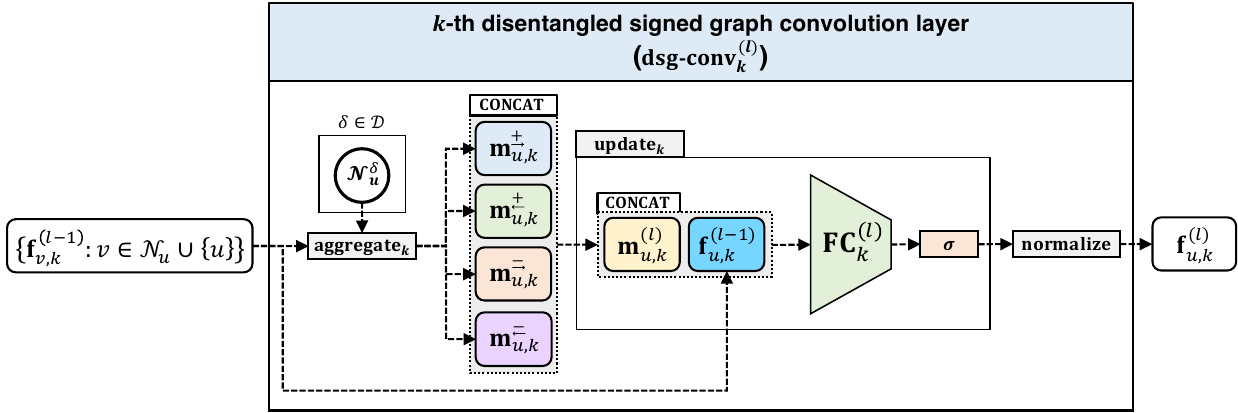}
    	\label{fig:dsg:conv}
    }
    \caption{
        Concept of disentangled signed graph convolution (\ourconv) at the $l$-th layer. 
    }
    \label{fig:dsg}
\end{figure}

For each factor $k$, the $\texttt{aggregate}_{k}$ function gathers the information of neighboring factors of node $u$ with neighbor type $\delta \in \mathcal{D}$, and the disentangled message $\msg{u}{k}{(l)}$ is generated by concatenating all the aggregated results.
The $\texttt{update}_k$ function performs a non-linear transformation from the previous factor $\chn{u}{k}{l-1}$ and the message $\msg{u}{k}{(l)}$, leading to the updated factor $\chn{u}{k}{l}$.
Specifically, their concatenation is fed into a fully-connected $\texttt{FC}$ layer regarding factor $k$ with a non-linear activation $\sigma$ where the concatenation with $\chn{u}{k}{l-1}$ is considered as a simple form of skip connection~\cite{Hamilton2017-bh} between the different layers.
After then, it normalizes the resulting factor for numerical stability, as described above.

For $\texttt{aggregate}(\cdot)$ in Equation~\eqref{eq:dsg:aggregate}, as opposed to previous approaches~\cite{Derr2018-bs, Li2020-py, shu2021sgcl, Huang2021-wp, Jung2022-pm}, we aim to aggregate the neighborhood without depending on the assumptions from the structural balance or status theory. 
Thus, we adopt and investigate the following simple aggregators that gather the information of $\delta$-neighbors where the type $\delta$ of neighbors is determined by only sign and direction (see Section~\ref{sec:prelim:notations}). 
\begin{itemize}[leftmargin=*]
    \item {\textbf{Sum aggregator}. 
        This simply adds the set of the $k$-th factor vectors of neighbors of node $u$ without any normalization as follows:
        \begin{align*}
            \msg{u}{k}{\delta}
            \defeq 
            \sum_{v \in \neighD_u}{\chnn{v}{k}}
        \end{align*}
        
        According to the previous study~\cite{Xu2018-sy}, the sum aggregator can provide a message that fully captures the local neighborhood structure as it is an injective function that distinctly maps a given multiset of features to a distinct output (or message).
    }
    \item {\textbf{Mean aggregator}. 
        This calculates the element-wise mean of the set of the $k$-th factor vectors by dividing the number $\card{\neighD_{u}}$ of neighbors as follows: 
        \begin{align*}
            \msg{u}{k}{\delta} 
            \defeq 
            \sum_{v \in \neighD_{u}}{
                \frac{1}{\card{\neighD_{u}}} \cdot \chnn{v}{k}
            }
        \end{align*}
        This approach is comparable to the convolutional rule employed in GCN~\cite{KipfW17}, emphasizing the significance of normalization in constraining the magnitude of messages.
        The mean aggregator is characterized by its ability to capture the distribution of features within a given set by taking the average of individual element features~\cite{Xu2018-sy}. 
    }
    \item {\textbf{Attention aggregator}. 
        Unlike the mean aggregator which gives an equal weight to all neighbors, the attention aggregator assigns different weights to the neighbors and takes the weighted average as follows:
        \begin{align*}
            \msg{u}{k}{\delta} 
            \defeq 
            \sum_{v \in \neighD_u}{
            \alpha_{k}^{\delta}(u,v) \cdot
            \chnn{v}{k}}
        \end{align*}
        where $v$ is a $\delta$-neighbor of node $u$, and $\alpha_{k}^{\delta}(u,v)$ is their attention probability w.r.t. the $k$-th factor, i.e., 

        \vspace{-3mm}
        \begin{align*}
        \alpha_{k}^{\delta}(u,v) = 
        \frac
            {\texttt{exp} \Big(
                \texttt{LeakyReLU}
                \big(
                \innerproduct{\vect{a }_{k}^{\delta}}
                {[\chnn{u}{k}{} \parallel \chnn{v}{k}]}
                \big) 
                \Big)}
            {\sum_{w \in \neighD_u}
                \texttt{exp} \Big(\texttt{LeakyReLU}
                \big(
                \innerproduct{\vect{a}_{k}^{\delta}}
                {[\chnn{u}{k}{} \parallel \chnn{w}{k}]}
                \big) \Big)}
        \end{align*}
        where 
        $\innerproduct{\cdot}{\cdot}$ is an inner product of two vectors, and 
        $\vect{a}_{k}^{\delta}$ is a trainable parameter vector of size $\frac{2d}{K}$.
    }
    \item {\textbf{Max pooling aggregator}. 
        The max pooling aggregator {\color{red} }takes element-wise maximum values of the set of factors as follows:
        \begin{align*}
            \msg{u}{k}{\delta} 
            \defeq
            \texttt{max}
            \big(
            \set{
                \mat{W}_{k}^{\delta} \cdot \chnn{v}{k} :v \in \neighD_{u}
            }
            \big)
        \end{align*}
        where $\texttt{max}(\cdot)$ is the element-wise max operator and $\mat{W}_{k}^{\delta} \in \mathbb{R}^{\frac{d}{K} \times \frac{d}{K}}$ is a trainable parameter matrix.
        The max aggregator aims to effectively capture representative elements of the neighboring set~\cite{Hamilton2017-bh,Xu2018-sy} as it takes the important (or maximum) information for each element.
    }    
\end{itemize}

We stack $L$ layers of the \ourconv operation for each factor $k$ (lines~\ref{alg:encoder:agg:start}$-$\ref{alg:encoder:agg:end}), and set $\chn{u}{k}{L}$ to $\vect{z}_{u, k}$ as the final disentangled factor where we set $d_{L}$ to $\dout$ (line~\ref{alg:encoder:set}).
The disentangled representation of node $u$ is expressed as a list of $\set{\vect{z}_{u, k}}$  (line~\ref{alg:encoder:return}), i.e.,  $\vect{z}_{u} = (\vect{z}_{u, 1}, \cdots, \vect{z}_{u,K})$.
We apply our encoder to all nodes in order to obtain their final representations.

\subsection{Pairwise Correlation Decoder for Sign Prediction}
\label{sec:proposed:decoder}

As the sentiment of an edge can be affected by correlations between factors (see Figure~\ref{fig:motivation:factored}), 
we propose a new decoding strategy that explicitly models pairwise correlations of disentangled factors and uses them as the edge's feature for the link sign prediction task. 
The procedure of our decoder is described in Algorithm~\ref{alg:decoder}.

\smallsection{Pairwise correlation decoder (lines~\ref{alg:decoder:for:start}$-$\ref{alg:decoder:for:end})}
Given the set $\set{\vect{z}_{u}}$ of disentangled representations and the set $\edges$ of directed edges, our decoder predicts the probability $p_{uv}$ of the sign of each input edge $u \rightarrow v$.
For a directed edge $u \rightarrow v$ to be predicted, our decoder first reshapes $\vect{z}_{u}$ and $\vect{z}_{v}$ as follows:
\begin{align*}
	\mat{Z}_{u} = \texttt{CONCAT}(\vect{z}_u) = 
		\begin{bmatrix}
 			\vect{z}_{u, 1} & \cdots & \vect{z}_{u, K}
		\end{bmatrix}
	\qquad
	\mat{Z}_{v} = \texttt{CONCAT}(\vect{z}_v) = 
		\begin{bmatrix}
 			\vect{z}_{v, 1} & \cdots & \vect{z}_{v, K}
		\end{bmatrix}
\end{align*}
where $\mat{Z}_{u} \in \mathbb{R}^{\frac{\dout}{K} \times K}$ is a matrix horizontally concatenating the column vectors $\vect{z}_{u,k} \in \mathbb{R}^{\frac{\dout}{K} \times 1}$ from $k=1$ to $K$.
The dimension of $\mat{Z}_{v}$ is the same as that of $\mat{Z}_{u}$.
We then obtain pairwise correlations between factors as follows:
\begin{align}
	\mat{H}_{uv} = \matt{Z}_{u}\mat{Z}_{v} = 
	\begin{bmatrix}
		\ip{1}{1} & \cdots & \ip{1}{K} \\
		\vdots & \ddots & \vdots \\
		\ip{K}{1} & \cdots & \ip{K}{K} \\
	\end{bmatrix}
\end{align}
where $\mat{H}_{uv} \in \mathbb{R}^{K \times K}$ is a factor-to-factor correlation matrix, i.e., the inner product $\innerproduct{\vect{z}_{u,i}}{\vect{z}_{v,j}}$ is a scalar interpreted as a correlation between $i$-th factor of node $u$ and $j$-th factor of node $v$ for $1 \leq i, j \leq K$. 
We call $\mat{H}_{uv}$ the edge feature map of edge $u \rightarrow v$.

 \begin{algorithm}[t!]
     \caption{Decoder of \method for link sign prediction}
     \label{alg:decoder}
     \begin{algorithmic}[1]
     	\small
         \Require 
         \Statex Set of disentangled node representations: $\set{\vect{z}_{u}$ : $u \in \mathcal{V} }$
         \Statex Set of directed edges: $\edges$
         \Ensure
         \Statex Set of predicted probabilities: $\set{p_{uv} : (u \rightarrow v) \in \edges}$
         \For{\textbf{each} $(u \rightarrow v) \in \edges$}\label{alg:decoder:for:start}
             \State $\mat{Z}_{u} \leftarrow \texttt{CONCAT}(\vect{z}_{u})$ and $\mat{Z}_{v} \leftarrow \texttt{CONCAT}(\vect{z}_{v})$
             \label{alg:decoder:concat}
             \State $\mat{H}_{uv} \leftarrow \matt{Z}_{u} \mat{Z}_{v}$ 
             {\color{cobalt}\Comment{\textbf{\footnotesize generate the edge feature map of $u \rightarrow v$}}}
             \label{alg:decoder:corr}
             \State $p_{uv} \leftarrow \texttt{sigmoid}\big(\texttt{sum}\left(\mat{H}_{uv} \odot \mat{W}_{s}\right)\big)$ \label{alg:decoder:prob}
             {\color{cobalt}\Comment{\textbf{\footnotesize predict the probability of a sign}}}
         \EndFor\label{alg:decoder:for:end}
         \State \textbf{return} $\set{p_{uv} : (u \rightarrow v) \in \edges}$         \label{alg:decoder:return}
     \end{algorithmic}
 \end{algorithm}

After then, our decoder calculates a score (or logit) $s_{uv}$ given $\mat{H}_{uv}$ as the following:
\begin{align}
	\label{eq:decoder:logit}
	s_{uv} = \texttt{sum}(\mat{H}_{uv} \odot \mat{W}_{s}) = 
	\sum_{i=1}^{K}\sum_{j=1}^{K} w_{ij} \ip{i}{j}
\end{align}
where $\texttt{sum}(\cdot)$ is a function summing all entries of an input matrix, $\mat{W}_{s} \in \mathbb{R}^{K \times K}$ is a learnable weight matrix, $\odot$ is the element-wise multiplication (or the Hadamard product), and $w_{ij}$ is $(i,j)$-th weight of $\mat{W}_{s}$.
Note that our decoder learns $O(K^2)$ correlations with weights $\set{w_{ij}}$ as shown in Equation~\eqref{eq:decoder:logit}.
Finally, the decoder computes the prediction probability as $p_{uv} = \texttt{sigmoid}(s_{uv})$ for the edge (line~\ref{alg:decoder:prob}), and returns the set of predicted probabilities of all given edges (line~\ref{alg:decoder:return}).

\smallsection{Loss function for link sign prediction}
The target task is considered a binary classification problem because it has two $+$ and $-$ classes. Thus, after computing the probability $p_{uv}$, \method optimizes the following standard binary cross entropy (\texttt{bce}) loss during training:
\begin{align}
	\label{eq:bce}
	\mathcal{L}_{\texttt{bce}} = -\frac{1}{\card{\edges}}\!\sum_{(u \rightarrow v) \in \edges}\!\!\big(y_{uv} \log{(p_{uv})} + \left(1 - y_{uv}\right)\log{\left(1 - p_{uv}\right)}\big)
\end{align}
where $\edges$ is the set of training edges, and $y_{uv}$ is the ground-truth of the sign of $u \rightarrow v$, i.e., $y_{uv} = 1$ for $+$ sign, or $y_{uv} = 0$ for $-$ sign.
After the training phase of \method is complete, the classifier predicts the sign of a test edge $u \rightarrow v$ as $+$ sign if $p_{uv} \geq 0.5$; otherwise, it is classified as $-$ sign.

\smallsection{Comparative Analysis}
The conventional approach in an entangled framework for building the edge's feature of two nodes $u$ and $v$ is to simply concatenate their corresponding embeddings~\cite{Derr2018-bs,Li2020-py,Liu2021-np,shu2021sgcl,Jung2022-pm,Yan2023-ta}.
Similarly, we could have followed the existing concatenation strategy, but such an approach is limited in fully exploiting the advantage of our disentangled representations.  
For example, suppose the edge feature is made by the naive concatenation of $\vect{z}_{u}$ and $\vect{z}_{v}$, i.e., $\vect{h}_{uv} = \texttt{CONCAT}(\set{\vect{z}_{u}, \vect{z}_{v}}) \in \mathbb{R}^{2\dout}$.
Then, the score $s_{uv}$ is obtained with a trainable weight vector $\vect{w} \in \mathbb{R}^{2\dout}$ as follows:
\begin{align}
	\label{eq:decoder:naive}
	s_{uv} = \innerproduct{\vect{w}}{\vect{h}_{uv}} = \sum_{k=1}^{K} \innerproduct{\vect{w}_{u,k}}{\vect{z}_{u,k}} + \sum_{k=1}^{K} \innerproduct{\vect{w}_{v,k}}{\vect{z}_{v,k}}
\end{align} 
where $\vect{w}_{u,k} \in \mathbb{R}^{\frac{\dout}{K}}$ is the sub-vector of parameters corresponding to $\vect{z}_{u,k}$ in $\vect{w}$. 
Equation~\eqref{eq:decoder:naive} implies that the concatenation strategy focuses on the information from each factor as $\innerproduct{\vect{w}_{u,k}}{\vect{z}_{u,k}}$, whereas our decoding strategy learns about correlations between factors as $w_{ij} \ip{i}{j}$ according to Equation~\eqref{eq:decoder:logit}.
Hence, our disentangled approach allows our model to leverage $K^2$ signals from the pairwise correlations of factors, which is more diverse than the naive concatenation strategy that utilizes only $2K$ signals from individual factors.
We empirically demonstrate that our decoding strategy is more effective than the traditional one in Section~\ref{sec:exp:ablation} (see Table~\ref{tab:ablation_study}).

\subsection{Self-supervised Factor-wise Classification}
\label{sec:proposed:ssl}
Although our encoder and decoder provide a framework that learns about disentangled factors and their correlations, this framework could be underperformed if we focus on only disentangling the embeddings, as in other disentangled GNNs~\cite{Ma2019-ce}.
Previous work~\cite{Liu2020-yd,zhao2022exploring} claimed that such GNNs could fail to sufficiently diversify latent factors as the relationship between the factors is likely to be complicated. 
Thus, it is desirable to encourage the disentanglement between factors under the hypothesis that the representations of latent factors should be distinct. 

In this work, we also follow the suggested direction for enhanced disentanglement. 
Specifically, we adopt a self-supervised approach utilized in~\cite{zhao2022exploring}, and investigate its effect on our purpose.
The predefined (or pretext) task is to classify each factor of a node so that it aims to boost the difference across factors (i.e., well-disentangled factors should be distinguishable from each other). 
For each node $u$, we generate a pseudo label $y_{u,k}$ which is set to the factor index of $\vect{z}_{u,k}$, i.e., $y_{u,k} = k$ where $1 \leq k \leq K$.
Then, we train a classifier for the factor-wise classification task by optimizing the following discriminative loss:
\begin{align}
	\label{eq:disc}
	\mathcal{L}_{\text{disc}} = -\frac{1}{nK}\sum_{u \in \vertex}\sum_{k=1}^{K}\sum_{i=1}^{K}\mathbb{I}(y_{u,k} = i)\log{\Big(\texttt{softmax}_{i}\big(\texttt{FC}_{\texttt{disc}}(\vect{z}_{u,k})\big)\Big)}
\end{align}
where $\vertex$ is the set of nodes, $n$ is the number of nodes, $K$ is the number of factors, $\texttt{softmax}_{i}(\cdot)$ is $i$-th probability after the softmax operation, and $\mathbb{I}(\cdot)$ is an indicator function that returns $1$ if a given predicate is true, and $0$ otherwise. 
$\texttt{FC}_{\texttt{disc}}(\cdot)$ consists of a weight matrix $\mat{W}_{\texttt{disc}} \in \mathbb{R}^{K \times \frac{d_{L}}{K}}$ and a bias vector $\vect{b}_{\texttt{disc}} \in \mathbb{R}^{K}$.

\subsection{Final Loss Function} 
We jointly train the above discriminator on top of our encoder and decoder in an end-to-end scheme by optimizing the following loss function:
\begin{align}
	\mathcal{L} = \mathcal{L}_{\texttt{bce}} + \lambda_{\texttt{disc}}\mathcal{L}_{\texttt{disc}} + \lambda_{\texttt{reg}}\mathcal{L}_{\texttt{reg}}
\end{align}
where $\mathcal{L}_{\texttt{bce}}$ is the cross entropy loss of Equation~\eqref{eq:bce} of the decoder connected from the encoder, and \lossssl is the discriminative loss of Equation~\eqref{eq:disc} with its strength \lambdassl.
We further optimize the regularization loss $\mathcal{L}_{\texttt{reg}}$ based on $\ell_{2}$-regularization (or weight decay) of model parameters to control overfitting where $\lambda_{\texttt{reg}}$ is the strength of regularization.

\subsection{Computational Complexity Analysis}
\label{sec:proposed:complexity}

We analyze the computational complexity of \method in this section. 
In the following analysis, $n=\card{\vertex}$ and $m=\card{\edges}$ are the numbers of nodes and edges, respectively. 
Below, $K$ is the number of factors, $L$ is the number of layers, and $d_{*}$ is the size of dimensions where $*$ is a wildcard character.

\begin{theorem}[Time Complexity]
\label{theorem:time}
	The time complexity of \method is $O(m+n)$ where $L$, $K$, and $d$ are fixed constants, assuming all of $d_{*}$ are set to $d$. 
\begin{proof*}
	The time complexity of each step in \method is summarized in Table~\ref{tab:time_complexities} at Appendix~\ref{appendix:analysis:time}.
	The time complexities of our encoder and decoder are proved in Lemmas~\ref{lemma:encoder}~and~\ref{lemma:decoder}, respectively. 
	It takes $O(m)$ to compute the cross entropy loss $\loss_{\texttt{bce}}$. 
	For computing \lossssl, $\texttt{FC}_{\texttt{disc}}(\cdot)$ takes $O(d + K)$ time, and $\texttt{softmax}(\cdot)$ takes $O(K)$ time. 
	We only consider the case when the indicator function is true; thus, the total cost of \lossssl for all nodes and $k$ is $O(dKn + K^2n)=O(dKn)$ where $K \leq d$.
    Therefore, the total time complexity is $O(m + n)$.
\end{proof*}
\end{theorem}

Theorem~\ref{theorem:time} indicates that \method is linearly scalable w.r.t. the number $m$ of edges where $n \leq m$ in real-world graphs as shown in Table~\ref{tab:data}.

\section{Experiments}
\label{sec:experiments}

In this section, we empirically analyze the effectiveness of our proposed \method for the link sign prediction task on real-world signed graphs. 
We aim to answer the following questions:

\begin{itemize}[leftmargin=*]
    \setlength\itemsep{0.5em}
    \item {
        \textbf{Q1. Accuracy~(Section~\ref{sec:exp:auccracy}).} 
        How accurately does \method predict the sign of a link compared to other state-of-the-art methods?
    }
    \item{
        \textbf{Q2. Effect of aggregators~(Section~\ref{sec:exp:aggregator}).}
        How does each aggregator for the signed graph convolution affect the accuracy of \method?
    }
    \item{
        \textbf{Q3. Ablation study~(Section~\ref{sec:exp:ablation}).}
        How does each module of \method affect its performance in the link sign prediction task?
    }
    \item{
        \textbf{Q4. Effect of hyperparameters~(Section~\ref{sec:exp:effect}).}
        How do \method's hyperparameters affect its performance in the link sign prediction task?
    }
    \item{
        \textbf{Q5. Qualitative study~(Section~\ref{sec:exp:qualitative}).}
        Does the discriminative loss of \method make factors more distinguishable in their embedding space? 
    }
    \item {
        \textbf{Q6. Computational efficiency~(Section~\ref{sec:exp:scalability}).} 
        Is our \method linearly scalable w.r.t. the number of edges? 
        Does our method provide a good trade-off between accuracy and training time?
    }
\end{itemize}

\def\arraystretch{1.2} 
\begin{table}[!t]
\begin{center}
\begin{minipage}{\textwidth}
\small
\caption{
Data statistics of signed graphs. 
$\card{\mathcal{V}}$ and $\card{\mathcal{E}}$ indicate the number of nodes and edges, respectively. 
$\card{\mathcal{E}^{s}}$ is the number of edges having $s$ sign where $s$ is $+$ or $-$.
$\rho(+)$ is the ratio of positive edges, i.e., $\rho(+)=\card{\mathcal{E}^{+}}/\card{\mathcal{E}}$.
}
\label{tab:data}
\begin{tabular*}{\textwidth}{@{\extracolsep{\fill}}lrrrrr@{\extracolsep{\fill}}}
    \hline\toprule
    \textbf{Dataset} & $\card{\mathcal{V}}$ & $\card{\mathcal{E}}$ & $\card{\mathcal{E}^{+}}$ & $\card{\mathcal{E}^{-}}$ & $\rho(+)$ \\ 
    \midrule
    BC-Alpha & 3,783 & 24,186 & 22,650 & 1,536 & 93.6 \\
    BC-OTC & 5,881 & 35,592 & 32,029 & 3,563 & 90.0 \\
    Wiki-RFA & 11,258 & 178,096 & 138,473 & 38,623 & 78.3 \\
    Slashdot & 79,120 & 515,397 & 392,326 & 123,255 & 76.1 \\
    Epinions & 131,828 & 841,372 & 717,667 & 123,705 & 85.3 \\ 
    \bottomrule\hline
\end{tabular*}   
\end{minipage}
\end{center}
\end{table}

\subsection{Experimental Setting}
\label{sec:exp:setting}

We describe our experimental setting including datasets, competitors, training details, and hyperparameter settings.

\smallsection{Datasets} 
We conduct experiments on five signed social networks whose statistics are summarized in Table~\ref{tab:data}.
BC-Alpha and BC-OTC~\cite{Kumar2016-lm} are who-trusts-whom networks of users on Bitcoin platforms.
Wiki-RFA~\cite{West2014-ij} is a signed network in Wikipedia where users vote for administrator candidates.
Slashdot~\cite{Kunegis2009-hw} is a signed social network on Slashdot, a technology news website where a user can tag other users as friends or foes.
Epinions~\cite{Guha2004-yz} is an online social network on Epinions, a general consumer review site where a user can express trustworthiness on others based on their reviews.
Note that all of the datasets are signed directed graphs. 
We provide more details of the datasets in Appendix~\ref{appendix:exp:data}.

\smallsection{Competitors}
We extensively compare our proposed method \method to the following competitors.
\begin{itemize}[leftmargin=*]
    \item {
        \textit{Signed network embedding methods}: SNE~\cite{yuan2017sne}, SIDE~\cite{Kim2018-aa}, BESIDE~\cite{ChenQLS18}, SLF~\cite{xu2019link}, ASiNE~\cite{lee2020asine}, and DDRE~\cite{xu2022dual}.
    }
    \item{
        \textit{Disentangled unsigned GNN}: DisenGCN~\cite{Ma2019-ce}.
    }
    \item{
        \textit{Signed GNNs}: SGCN~\cite{Derr2018-bs}, SNEA~\cite{Li2020-py}, SGCL~\cite{shu2021sgcl}, GS-GNN~\cite{Liu2021-np}, SDGNN \cite{Huang2021-wp}, SIDNET~\cite{Jung2022-pm}, MUSE~\cite{Yan2023-ta} and SigMaNet~\cite{Fiorini2022-ef}.
    }
\end{itemize}

We provide detailed information regarding the implementation of each competitor in Appendix~\ref{appendix:url}.
We exclude unsigned network embedding methods such as Deepwalk~\cite{Perozzi2014-qc}, LINE~\cite{Tang2015-xs}, and Node2vec~\cite{Grover2016-qz} since previous studies~\cite{Huang2021-wp,ChenQLS18,yuan2017sne} have demonstrated that these methods are inferior to the methods using the sign information in the link sign prediction task. 

Note that the original datasets are represented as signed directed graphs, and we aim to learn signed directed graphs without any modification to the datasets.
Following \cite{Huang2021-wp,Jung2022-pm}, we conduct experiments on all methods including ASiNE, SGCN, SNEA, and GS-GNN in signed directed graphs formed from the non-filtered raw data.
For DisenGCN, we use only edges without signs for training because the model takes an unsigned graph as input.

\smallsection{Data split and evaluation metrics}
We perform link sign prediction (i.e., predicting whether links are positive or negative), a standard benchmark task in signed graphs, to evaluate the performance of each model.
For the experiments, we randomly split the edges by the 8:2 ratio into training and test sets.
Note that the ratio of signed edges is skewed significantly towards positive edges as shown in Table~\ref{tab:data}.
Considering the imbalance, we use AUC~\cite{Bradley1997-oi} and Macro-F1~\cite{Taha2015-oi} in percentage, as metrics of the task.
For each metric, we repeat experiments $10$ times with different random seeds and report average test results with standard deviations.

\smallsection{Implementation details}
We utilize the Adam optimizer for training where $\eta$ is its learning rate.
We implement all models including our \method in Python 3.9 using PyTorch 1.10. 
We use a workstation with Intel Xeon Silver 4215R and Geforce RTX 3090 (24GB VRAM).

\smallsection{Hyperparameter settings}
For all tested models, we set the dimension of the final node embeddings to $64$ so that the embeddings have an equal capacity for learning the downstream task.
We also fix the number of epochs to $100$ so that each model reads the same amount of data for training.
For each method, we use $5$-fold cross-validation to search for the best hyperparameters and compute the test accuracy with the validated ones. 
The search space of each hyperparameter of \method is as follows: 
$\set{4, 8, 16}$ for the number $K$ of factors, 
$\set{2, 4, 6, 8}$ for the number $L$ of layers,
$\set{0.1, 0.5, 1}$ for the strength \lambdassl of the discriminative loss, and
$\set{0.01, 0.005, 0.001}$ for the learning rate $\eta$ and the strength \lambdareg of regularization where validated results are reported in Appendix~\ref{appendix:hyperparam}.
We follow the range of each hyperparameter reported in its corresponding paper for other models.

\subsection{Performance of Link Sign Prediction}
\label{sec:exp:auccracy}

\smallsection{Predictive performance}
We evaluate all methods on real-world signed graphs for the link sign prediction task where the results are summarized in Tables~\ref{tab:sign:auc}~and~\ref{tab:sign:f1} in terms of AUC and Macro-F1, respectively. 
For \method, the sum aggregator is used for its optimal performance (see its effect in Table~\ref{tab:aggregator}).
We observe the following from the tables:
\begin{itemize}[leftmargin=*]
    \item{
        Our proposed \method shows the best performance among all tested methods in terms of both metrics across all datasets. 
        Compared to the second-best results, our approach improves AUC by up to 3.1\% and Macro-F1 by up to 6.5\%, respectively.
    }
    \item {
        The performance of DisenGCN, an unsigned GCN, is much worse than that of \method in most datasets. 
        This result implies that it is important to consider the information from signs even in the disentangled framework.
    }
    \item {
        \method outperforms methods relying on the balance or status theories (e.g., SIDE, BESIDE, SGCN, SNEA, SDGNN, and MUSE). 
        This result validates that node embeddings are effectively trained through our disentangled approach without any assumption from the theories.
    }
    \item {
        Note that the performance of our \method is significantly better than that of other models in terms of Macro-F1 (i.e., the mean of per-class F1 scores).
        This result implies \method is able to more accurately distinguish the sign of a given edge even for heavily imbalanced data as shown in Table~\ref{tab:data}.
    }
\end{itemize}

\begin{table}[!t]
\begin{center}
\begin{minipage}{\textwidth}
\small
\caption{
Our \method exhibits the best performance of link sign prediction in terms of AUC. 
The best result is marked in boldface, and the second-best result is underlined.
The \% diff means how much the best accuracy improved over the second-best accuracy.
The o.o.m. stands for out-of-memory error.
}
\label{tab:sign:auc}
\begin{tabular*}{\textwidth}{@{\extracolsep{\fill}}c@{}ccccc}
\hline\toprule
\textbf{AUC}          & \textbf{BC-Alpha}                  & \textbf{BC-OTC}                    & \textbf{Wiki-RFA}                  & \textbf{Slashdot}                  & \textbf{Epinions}                  \\ \midrule
SNE            & 79.4$\pm$1.5          & 82.8$\pm$1.8          & 70.8$\pm$1.5          & 69.8$\pm$1.4          & 82.4$\pm$2.0          \\
SIDE           & 81.8$\pm$1.0          & 83.9$\pm$0.6          & 73.3$\pm$1.0          & 80.4$\pm$0.3          & 91.0$\pm$0.3          \\
BESIDE         & 89.9$\pm$0.6          & 91.9$\pm$0.6          & \underline{90.3$\pm$0.2}    & \underline{90.0$\pm$0.1}    & 93.9$\pm$0.1          \\
SLF            & 86.5$\pm$0.9          & 88.2$\pm$1.1          & 89.9$\pm$0.3          & 88.6$\pm$0.1          & 90.9$\pm$0.2          \\
ASiNE          & 82.5$\pm$1.4          & 83.0$\pm$0.7          & 70.9$\pm$0.7          & o.o.m.            & o.o.m.            \\
DDRE           & 88.4$\pm$0.8          & 91.3$\pm$0.5          & 89.1$\pm$0.2          & 89.0$\pm$0.1          & \underline{94.0$\pm$0.1} \\ \midrule
DisenGCN       & 84.4$\pm$1.7          & 86.9$\pm$1.1          & 70.2$\pm$1.6          & 73.7$\pm$1.3          & 86.6$\pm$0.6          \\
SGCN           & 86.0$\pm$0.8          & 88.2$\pm$0.7          & 77.0$\pm$0.3          & 86.8$\pm$0.1          & 90.7$\pm$0.2          \\
SNEA           & \underline{90.1$\pm$0.7}    & 90.7$\pm$0.7          & 88.0$\pm$0.2          & 75.8$\pm$3.3          & 84.1$\pm$0.5          \\
SGCL           & 89.2$\pm$0.9          & 91.6$\pm$0.7          & 90.1$\pm$0.2          & 87.7$\pm$0.9          & 93.5$\pm$0.3          \\
GS-GNN         & 88.5$\pm$0.7          & 91.6$\pm$0.4          & 85.5$\pm$0.4          & 88.2$\pm$0.1          & 92.7$\pm$0.1          \\
SDGNN          & 89.2$\pm$0.6          & 91.5$\pm$0.5          & 89.3$\pm$0.3          & 88.2$\pm$0.2          & 92.9$\pm$0.8          \\
SIDNET         & 89.4$\pm$0.6          & 92.0$\pm$0.8          & 89.8$\pm$0.1          & 89.2$\pm$0.1          & \underline{94.0$\pm$0.2}    \\
MUSE           & 89.7$\pm$0.9          & \underline{92.2$\pm$0.8}    & 90.1$\pm$0.3            & o.o.m.            & o.o.m.            \\
SigMaNet       & 86.6$\pm$0.7          & 89.4$\pm$0.6          & 86.2$\pm$0.2          & 84.8$\pm$0.1          & 91.1$\pm$0.3          \\
\textbf{DINES (Ours)} & \textbf{92.8$\pm$0.8} & \textbf{95.1$\pm$0.7} & \textbf{91.3$\pm$0.3} & \textbf{92.7$\pm$0.2} & \textbf{96.7$\pm$0.1} \\ \midrule
\% diff        & 3.0\%             & 3.1\%             & 1.1\%             & 2.8\%             & 2.9\%            \\
\bottomrule\hline
\end{tabular*}
\end{minipage}
\end{center}
\end{table}

\begin{table}[!t]
\begin{center}
\begin{minipage}{\textwidth}
\small
\caption{
Our \method also shows the best performance in terms of Macro-F1.
}
\label{tab:sign:f1}
\begin{tabular*}{\textwidth}{@{\extracolsep{\fill}}c@{}ccccc}
\hline\toprule
\textbf{Macro-F1}     & \textbf{BC-Alpha}                  & \textbf{BC-OTC}                    & \textbf{Wiki-RFA}                  & \textbf{Slashdot}                  & \textbf{Epinions}                  \\ \midrule
SNE                   & 65.2$\pm$1.7                     & 72.2$\pm$1.9                   & 61.6$\pm$1.3                     & 62.2$\pm$1.0                     & 73.6$\pm$2.9                     \\
SIDE                  & 62.7$\pm$1.7                     & 69.1$\pm$1.1                   & 53.4$\pm$2.1                     & 68.1$\pm$1.7                     & 81.2$\pm$0.6                     \\
BESIDE                & 72.7$\pm$1.0                     & 79.0$\pm$1.0                   & 78.5$\pm$0.2                     & 79.1$\pm$0.1                     & 86.3$\pm$0.1                     \\
SLF                   & 74.0$\pm$1.2                     & 78.9$\pm$1.0                   & \underline{79.0$\pm$0.2}               & \underline{79.2$\pm$0.1}               & 84.0$\pm$0.6                     \\
ASiNE                 & 61.6$\pm$1.8                     & 67.0$\pm$1.5                   & 52.8$\pm$0.4                     & o.o.m.                       & o.o.m.                       \\
DDRE                  & 74.9$\pm$1.6                     & 79.1$\pm$0.6                   & 77.3$\pm$0.3                     & 77.8$\pm$0.0                     & 86.3$\pm$0.1                     \\ \midrule
DisenGCN              & 63.2$\pm$1.8                     & 73.1$\pm$1.8                   & 49.2$\pm$2.5                     & 63.6$\pm$4.2                     & 76.5$\pm$0.6                     \\
SGCN                  & 66.7$\pm$1.7                     & 73.5$\pm$1.9                   & 65.1$\pm$1.0                     & 76.0$\pm$0.3                     & 75.9$\pm$0.2                     \\
SNEA                  & 73.1$\pm$1.8                     & 77.3$\pm$1.3                   & 76.1$\pm$0.2                     & 43.2$\pm$0.0                     & 77.5$\pm$0.9                     \\
SGCL                  & 72.2$\pm$3.9                     & 76.0$\pm$1.3                   & 67.4$\pm$1.0                     & 67.3$\pm$2.5                     & 80.0$\pm$1.9                     \\
GS-GNN                & 73.7$\pm$1.6                     & 80.7$\pm$0.7                   & 71.5$\pm$1.9                     & 77.3$\pm$0.3                     & \underline{86.9$\pm$0.1}               \\
SDGNN                 & 73.9$\pm$1.5                     & \underline{80.8$\pm$0.7}             & 77.8$\pm$0.5                     & 77.1$\pm$0.4                     & 86.3$\pm$0.3                     \\
SIDNET                & \underline{75.2$\pm$1.5}               & 79.5$\pm$1.0                   & 77.7$\pm$0.6                     & 78.4$\pm$0.2                     & 85.8$\pm$0.2                     \\
MUSE                  & 73.9$\pm$1.5                     & 80.6$\pm$0.7                   & 78.6$\pm$0.2                       & o.o.m.                       & o.o.m.                       \\
SigMaNet              & 72.0$\pm$1.6                     & 77.8$\pm$0.9                   & 74.8$\pm$0.3                     & 74.4$\pm$0.2                     & 82.4$\pm$0.3                     \\
\textbf{DINES (Ours)} & \textbf{80.1$\pm$1.7}            & \textbf{85.3$\pm$1.0}          & \textbf{79.6$\pm$0.7}            & \textbf{82.7$\pm$0.6}            & \textbf{89.7$\pm$0.5}            \\ \midrule
\% diff               & 6.5\%                        & 5.5\%                      & 0.8\%                        & 4.0\%                        & 3.2\%  \\
\bottomrule\hline
\end{tabular*}
\end{minipage}
\end{center}
\end{table}

\begin{figure}[!t]
    \vspace{-5mm}
    \centering
    \includegraphics[width=\textwidth]{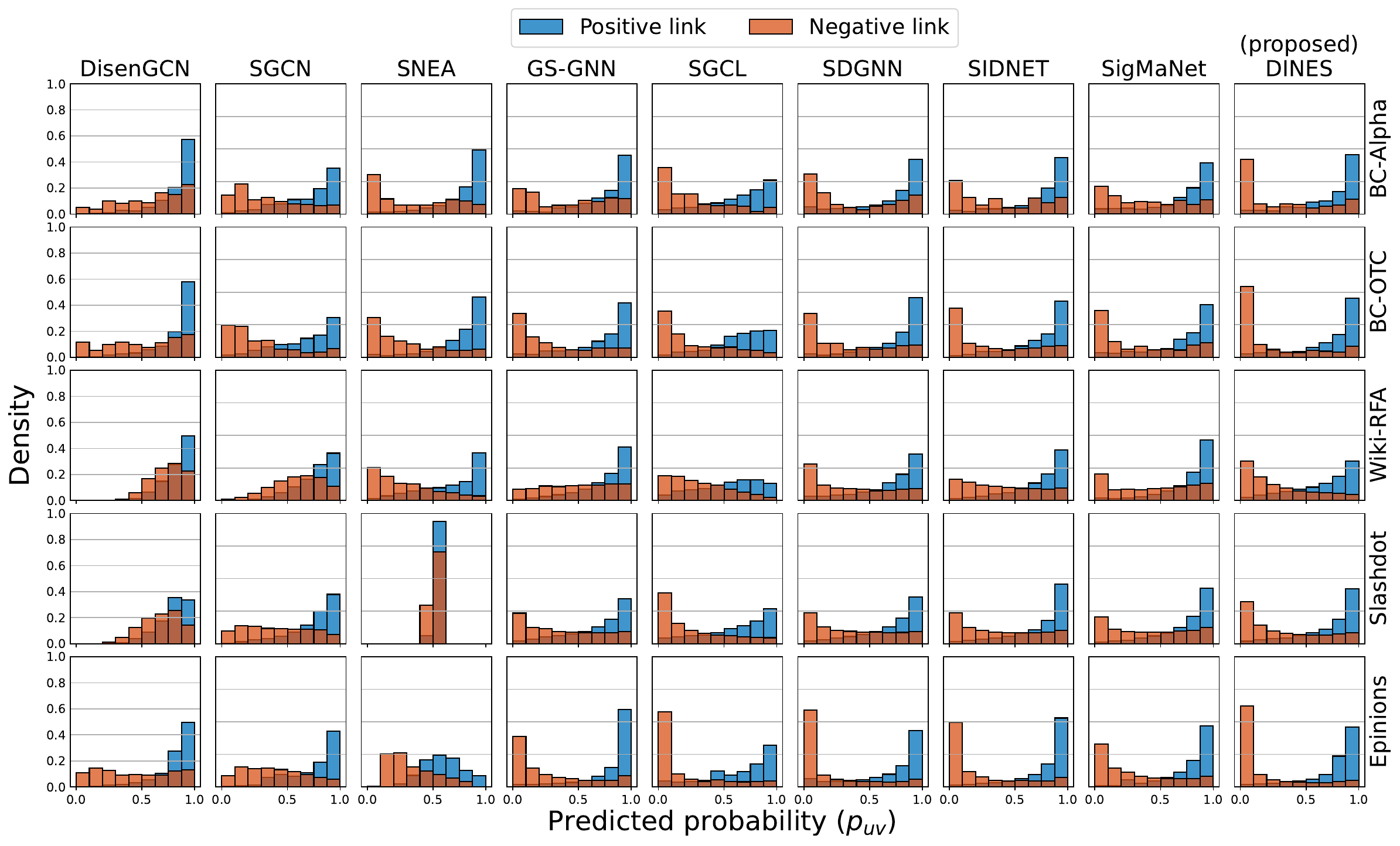}
    \caption{
        Distribution of predicted probabilities $p_{uv}$ of each model.  
        Our \method predicts the probabilities more confidently and precisely than other methods as the distributions of \method are more sharply U-shaped. 
    }
    \label{fig:experiments:edge_probs_hist}
\end{figure}

\smallsection{Distribution of predicted probabilities}
We further analyze the distribution of predicted probabilities $p_{uv}$ of each GNN model's classifier for given test edges $(u, v)$, as shown in Figure~\ref{fig:experiments:edge_probs_hist}, where MUSE is not considered due to its inability to handle large datasets.
Note that a desirable classifier should compute high (or low) scores for positive (or negative) edges so that the distribution for the positive (or negative) class should be left (or right) skewed, i.e., a U-shaped distribution is desirable.

As demonstrated in Figure~\ref{fig:experiments:edge_probs_hist}, the distributions of \method are more distinct than those of other methods, especially at the leftmost and the rightmost ends. 
Specifically, our method predicts negative signs more confidently as well as positive signs compared to other state-of-the-art methods such as GS-GNN, SDGNN, and SIDNET. 
Note that DisenGCN produces an undesirable distribution for negative edges compared to other signed GNNs, implying that it is crucial to use the sign information for learning.

\def\arraystretch{1.2} 
\setlength{\tabcolsep}{1pt} 
\begin{table}[!t]
\small
\begin{threeparttable}[t]
\caption{
Effect of different aggregators such as sum, max, mean, and attention on the performance of \method.
The sum aggregator consistently achieves the best accuracy across all datasets.
}
\label{tab:aggregator}
\centering
\begin{tabular*}{\textwidth}{@{\extracolsep{\fill}}clccccc}
\hline\toprule
\textbf{Metric}                    & \textbf{Variants} & \small{\textbf{BC-Alpha}} & \small{\textbf{BC-OTC}}   &  \small{\textbf{Wiki-RFA}}  & \small{\textbf{Slashdot}} & \small{\textbf{Epinions}} \\ \midrule
\multirow{4}{*}{\textbf{AUC}}      & \methodsum          & \textbf{92.8$\pm$0.8} & \textbf{95.1$\pm$0.7} & \textbf{91.3$\pm$0.3} & \textbf{92.7$\pm$0.2} & \textbf{96.7$\pm$0.1} \\
                                   & \methodmax           & 92.1$\pm$0.6          & 94.6$\pm$0.5          & 91.2$\pm$0.1          & 92.0$\pm$0.2          & 96.7$\pm$0.2          \\
                                   & \methodmean          & 91.8$\pm$0.8          & 94.4$\pm$0.6          & 90.3$\pm$0.2          & 91.1$\pm$0.3          & 96.4$\pm$0.2          \\
                                   & \methodattn          & 92.3$\pm$0.5          & 94.5$\pm$0.5          & 90.3$\pm$0.6          & 91.2$\pm$0.3          & 96.4$\pm$0.2          \\ \midrule
\multirow{4}{*}{\textbf{Macro-F1}} & \methodsum          & \textbf{80.1$\pm$1.7} & \textbf{85.3$\pm$1.0} & \textbf{79.6$\pm$0.7} & \textbf{82.7$\pm$0.6} & \textbf{89.7$\pm$0.5} \\
                                   & \methodmax           & 77.0$\pm$1.5          & 83.9$\pm$1.3          & 78.8$\pm$1.4          & 81.3$\pm$0.6          & 89.2$\pm$0.5          \\
                                   & \methodmean          & 76.7$\pm$1.8          & 83.5$\pm$0.9          & 77.7$\pm$1.2          & 79.9$\pm$0.8          & 88.7$\pm$0.5          \\
                                   & \methodattn          & 76.7$\pm$1.8          & 84.0$\pm$1.0          & 78.4$\pm$0.9          & 80.4$\pm$0.9          & 88.9$\pm$0.4          \\ \bottomrule\hline
\end{tabular*}
\end{threeparttable}
\end{table}

\subsection{Effect of Aggregators} 
\label{sec:exp:aggregator}
We examine the effect of aggregators such as sum, mean, max, and attention on \method in the link sign prediction task.
Table~\ref{tab:aggregator} demonstrates that the sum aggregator outperforms the others  across all datasets. 
The primary distinction lies in the normalization of gathered message features.
The mean, max, and attention aggregators apply their own normalization, whereas the sum aggregator does not. 
According to \cite{Xu2018-sy}, this allows the sum aggregator to fully capture the structural information from the local neighborhood of a node, which may be compressed or condensed by the normalization in mean or max functions. 
In other words, the sum aggregator is injective, producing distinct messages from a multiset of neighboring features, unlike the others.
Thus, the sum aggregator can offer more expressive features, particularly in capturing such a local structure.

Especially, the local neighborhood information of a node is highly related to the node's degree.
Based on our further investigation in Appendix~\ref{appendix:degree}, we found that the magnitude of a gathered message using the sum aggregator exhibits a nearly perfect positive correlation with the degree, unlike other aggregators.
This implies that the degree information is effectively distributed across the message feature of the sum aggregator such that the norm of the feature is proportional to the degree. 
As a result, the sum aggregator produces informative features that encode the degree information, enabling our model to effectively learn from the data and achieve superior performance. 
This result aligns with the in-depth analysis conducted in~\cite{Leskovec2010-vk}, which highlights node degrees as powerful features for the link sign prediction task.

\subsection{Ablation Study} \label{sec:exp:ablation}
We conduct an ablation study to verify the effect of each module in \method: disentanglement (\textsc{d}), pairwise correlation decoder (\textsc{p}), and self-supervised classification (\textsc{s}). 
For the study, we define the following variants of \method:

\contourlength{0.001em}
\begin{itemize}[leftmargin=*]
    \item {
        \contour{black}{\methodS} excludes the self-supervised factor classification from \method. 
    }
    \item {
        \contour{black}{\methodSF} excludes the pairwise decoding strategy from \methodS. 
        It simply concatenates the embeddings of two nodes for the edge feature, i.e., \texttt{CONCAT$(\{\vect{z}_{u}, \vect{z}_{v}\})$} for a given edge $u \rightarrow v$.
    }
    \item {
        \contour{black}{\methodSFD} excludes the disentanglement functionality from \methodSF. It is equivalent to \methodSF with $K=1$. 
    }
    \vspace{0.5em}
\end{itemize}

\def\arraystretch{1.2} 
\setlength{\tabcolsep}{2pt} 
\begin{table}[!t]
\small
\begin{threeparttable}[t]
\caption{
Result of the ablation study.
Combined with disentanglement (\textsc{d}), pairwise correlations (\textsc{p}), and self-supervised classification (\textsc{s}), \method provides the best accuracy for most datasets in terms of AUC and Macro-F1.
}
\label{tab:ablation_study}
\centering
\begin{tabular*}{\textwidth}{@{\extracolsep{\fill}}clccccc}
\hline\toprule
\textbf{Metric}                    & \textbf{Variants} & \textbf{BC-Alpha} & \textbf{BC-OTC}   & \textbf{Wiki-RFA}  & \textbf{Slashdot} & \textbf{Epinions} \\ \midrule
\multirow{4}{*}{\textbf{AUC}}      & \method          & \textbf{92.8$\pm$0.8} & \textbf{95.1$\pm$0.7} & \textbf{91.3$\pm$0.3} & \textbf{92.7$\pm$0.2} & \textbf{96.7$\pm$0.1} \\
                                   & \methodS         & 92.5$\pm$0.9          & 94.8$\pm$0.4          & 91.0$\pm$0.2          & 91.6$\pm$0.1          & 96.3$\pm$0.2          \\
                                   & \methodSF        & 90.7$\pm$0.5          & 92.6$\pm$0.5          & 90.9$\pm$0.1          & 90.7$\pm$0.1          & 95.6$\pm$0.1          \\
                                   & \methodSFD       & 89.8$\pm$0.6          & 92.3$\pm$0.5          & 90.5$\pm$0.1          & 90.1$\pm$0.1          & 95.1$\pm$0.2          \\\midrule
\multirow{4}{*}{\textbf{Macro-F1}} & \method          & \textbf{80.1$\pm$1.7} & 85.3$\pm$1.0 & \textbf{79.6$\pm$0.7} & \textbf{82.7$\pm$0.6} & \textbf{89.7$\pm$0.5} \\
                                   & \methodS         & 79.2$\pm$0.7          & \textbf{85.4$\pm$0.7}          & 79.0$\pm$1.1          & 81.5$\pm$1.1          & 88.7$\pm$0.6          \\
                                   & \methodSF        & 75.5$\pm$1.5          & 81.9$\pm$0.6          & 79.0$\pm$0.3          & 79.7$\pm$0.2          & 88.0$\pm$0.1          \\
                                   & \methodSFD       & 75.8$\pm$1.5          & 81.7$\pm$0.9          & 79.0$\pm$0.4          & 79.4$\pm$0.3          & 87.4$\pm$0.2                  \\ \bottomrule\hline
\end{tabular*}
\end{threeparttable}
\end{table}

Table~\ref{tab:ablation_study} shows the result of the ablation study in terms of AUC and Macro-F1 of the link sign prediction task.
Note that \method achieves the best performance when all the modules are combined. 
We have the detailed observations as follows:
\begin{itemize}[leftmargin=*]
    \item {
        \methodS slightly underperforms compared to \method for all datasets except the BC-OTC dataset, implying the enhanced disentanglement from the self-supervised factor classification is helpful for the performance. 
    }
    \item {
        Regarding the effect of our decoding strategy, \methodSF performs much worse than \methodS, especially in the BC-Alpha and BC-OTC datasets. 
        This result indicates that edge features based on our pairwise correlation are effective for the task, compared to the naive concatenation strategy. 
    }
    \item {
        The disentanglement framework is also beneficial as \methodSF performs better than \methodSFD in most datasets.  
    }
    \vspace{0.5em}
\end{itemize}

\subsection{Effect of Hyperparameters}
\label{sec:exp:effect}

\begin{figure}[!t]
    \centering
    \includegraphics[width=0.8\linewidth]{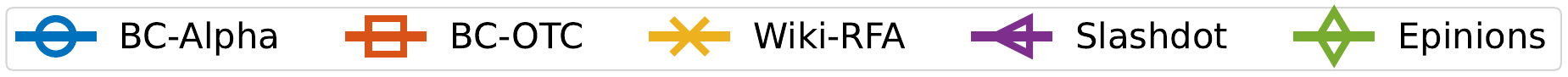}\vspace{-2mm}\\
    \subfigure[Number of factors]{
        \hspace{-5mm}
        \includegraphics[width=0.255\linewidth]{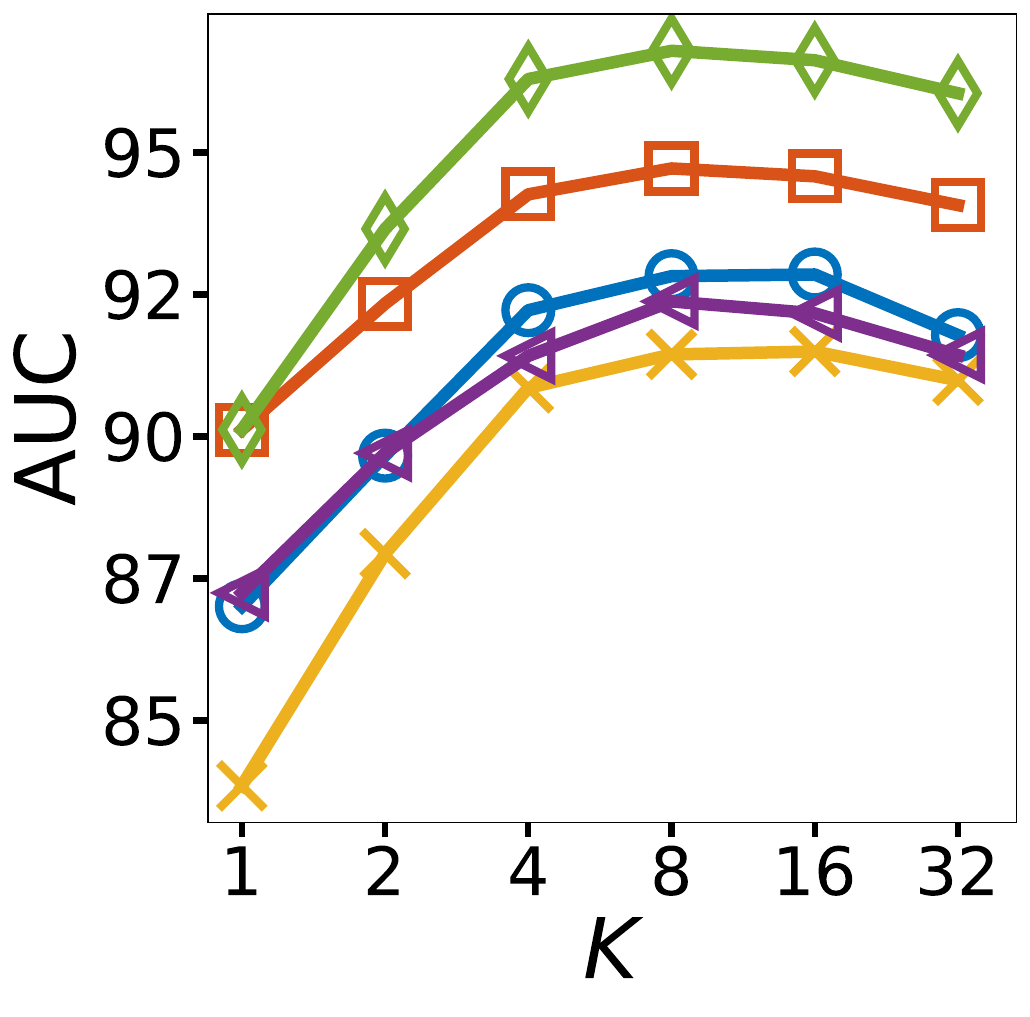}
        \hspace{-2mm}
        \label{fig:experiments:sensitivity:fac}
    }
    \subfigure[Number of layers]{
        \hspace{-2mm}
        \includegraphics[width=0.233\linewidth]{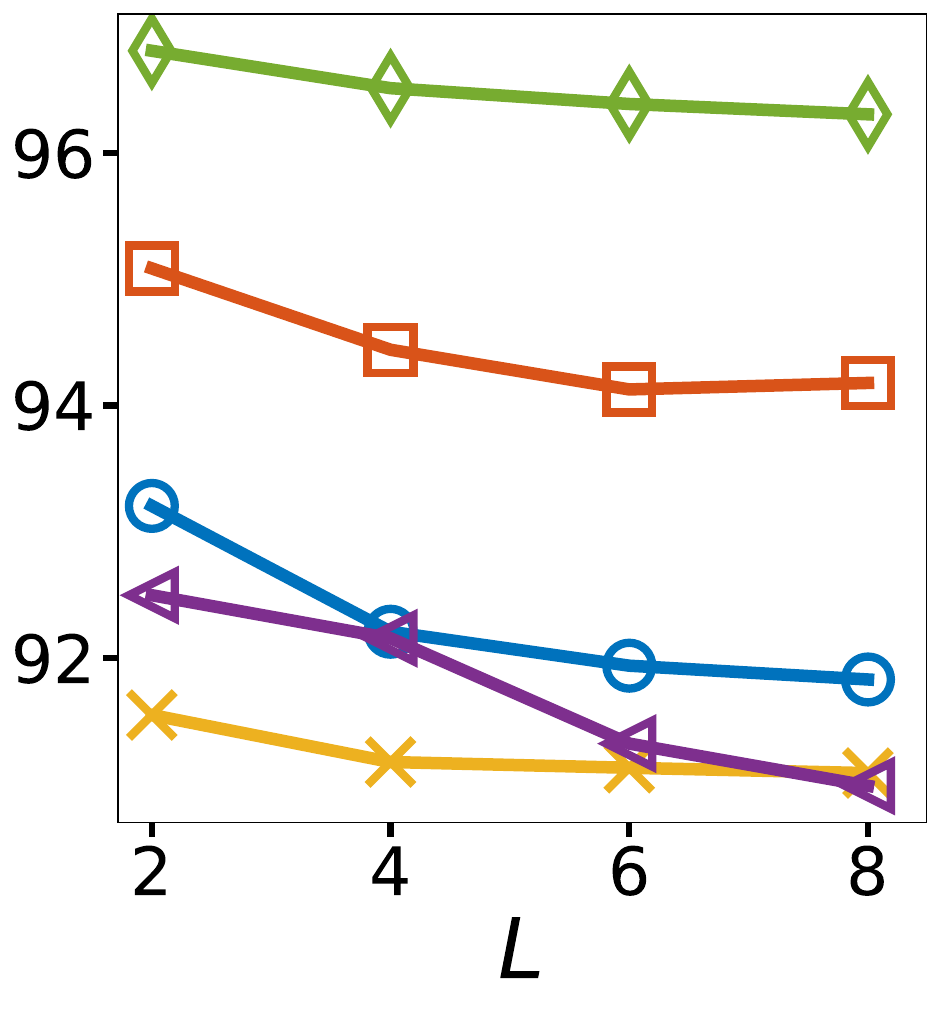}
        \hspace{-2mm}
        \label{fig:experiments:sensitivity:lay}
    }
    \subfigure[Discriminative loss]{
        \hspace{-2mm}
        \includegraphics[width=0.233\linewidth]{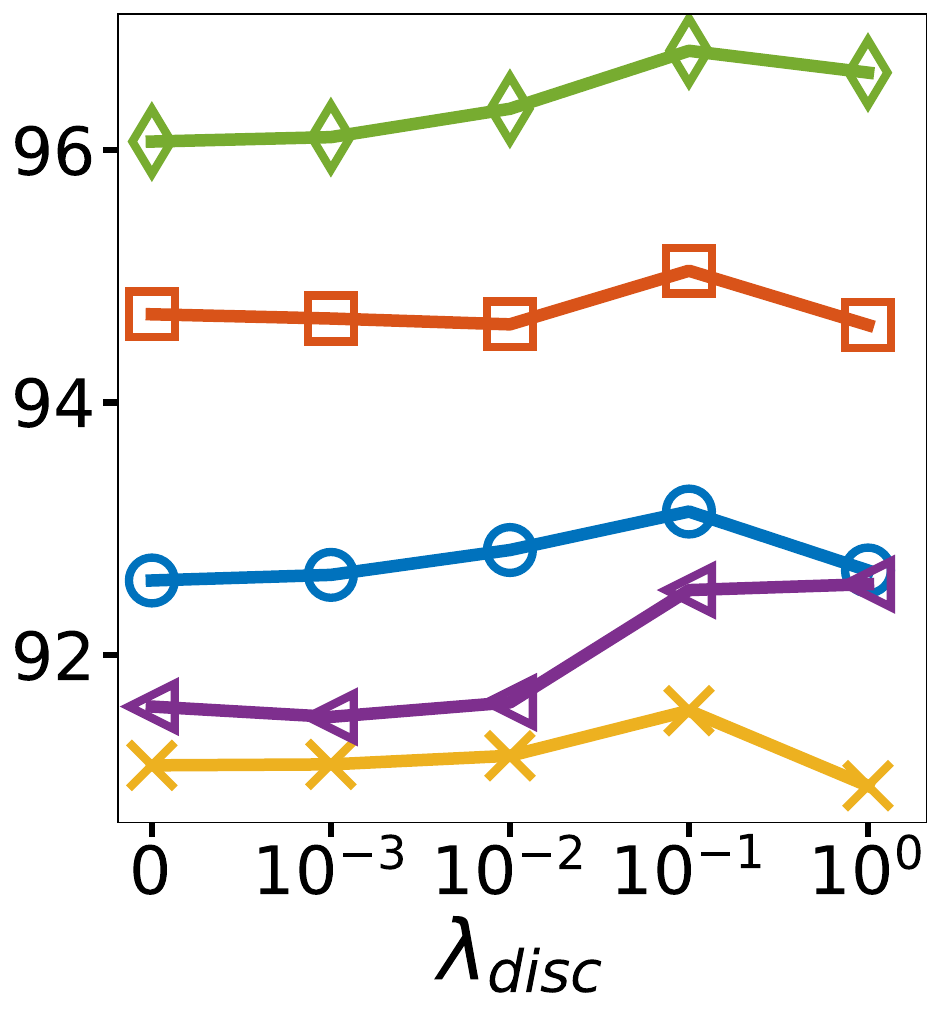}
        \hspace{-2mm}
        \label{fig:experiments:sensitivit:dis}
    }
    \subfigure[Regularization]{
        \hspace{-2mm}
        \includegraphics[width=0.233\linewidth]{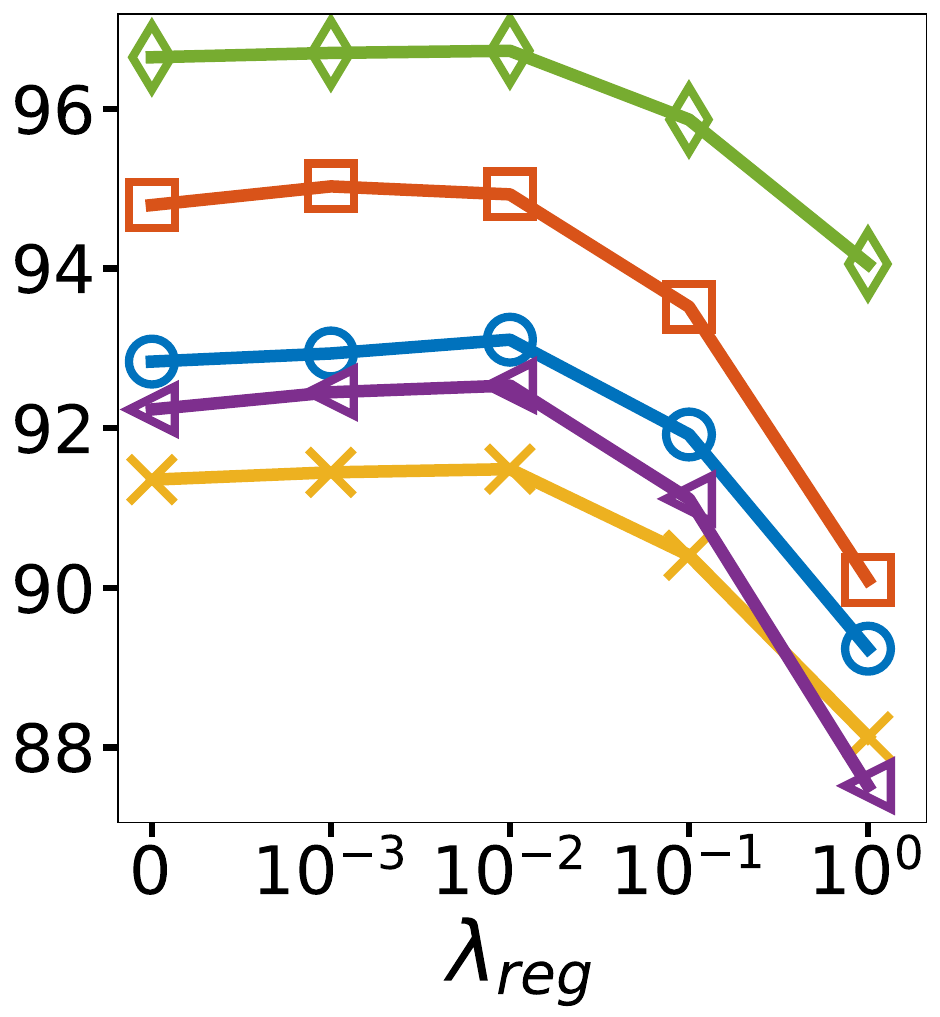}
        \label{fig:experiments:sensitivity:reg}
        \hspace{-4mm}
    }
    \caption{
        \label{fig:experiments:sensitivity}
        Effects of hyperparemters of \method where (a) $K$ is the number of factors, 
        (b) $L$ is the number of layers,
        (c) \lambdassl is the strength of the discriminative loss, and 
        (d) \lambdareg is the strength of regularization. 
    }
\end{figure}

We investigate the effects of hyperparameters $K$, $L$, \lambdassl, and \lambdareg of \method as shown in Figure~\ref{fig:experiments:sensitivity}.
For each hyperparameter, we vary its value while holding the other hyperparameters at the values obtained from the best settings. 
\begin{itemize}[leftmargin=*]
	\item {
		Figure~\ref{fig:experiments:sensitivity:fac} shows the effect of the number of factors, which is a crucial hyperparameter of \method as it controls the factors between nodes. 
		When $K=1$ and $2$, the performance is poor, while it tends to improve as the value of $K$ increases, implying that modeling latent and intricate relationships between nodes through the use of multiple factors is beneficial for enhancing performance. 
		However, the performance is degraded when $K=32$ because the dimension of each factor is too small.
		It's worth noting that this issue does not arise when the dimension is sufficient, as shown in Figure~\ref{fig:experiments:sensitivity:fac_dim}.
	}
	\item{
		Figure~\ref{fig:experiments:sensitivity:lay} demonstrates that \method generally performs better when the number of $L$ of layers is $2$.
	}
	\item{
		In general, too small or large values of \lambdassl are not a good choice because of their poor performances, as shown in Figure~\ref{fig:experiments:sensitivit:dis}.
	}
	\item{
		Figure~\ref{fig:experiments:sensitivity:reg} indicates that if the value of \lambdareg is greater than $0.01$, then it adversely affects the performance of \method as the model is heavily regularized.
	}
\end{itemize}

\begin{figure}[t]
    \centering
    \includegraphics[width=1.0\textwidth]{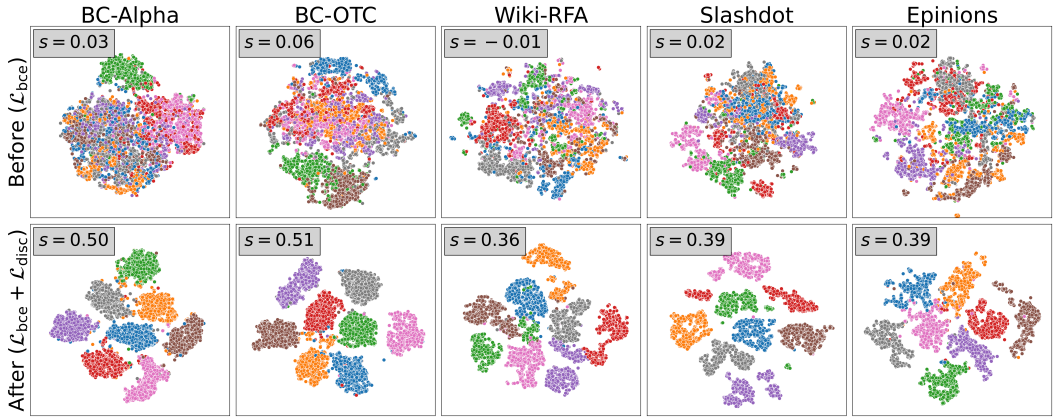}
    \caption{
        Visualization of disentangled node embeddings of \method before or after the discriminative loss \lossssl is applied (best viewed in color). 
    }
    \label{fig:experiments:visaulization}
\end{figure}

\subsection{Qualitative Analysis}
\label{sec:exp:qualitative}

We qualitatively analyze how the discriminative loss \lossssl of \method affects its disentangled representation learning.
For this purpose, we use t-SNE~\cite{van2008visualizing} to visualize disentangled node embeddings for each factor with a distinct color. 
Figure~\ref{fig:experiments:visaulization} demonstrates the visualization results for each dataset. 

As expected by the motivation of \lossssl, \method with the loss in the below sub-plots forms more distinct clusters for each factor on the projected space than that without the loss in the above sub-plots across all the datasets. 
The factors with the same class are agglomerated, and those between other classes are detached. 
We further numerically measure the cluster's cohesion for each factor in terms of silhouette coefficient~\cite{Rousseeuw1987-qe}, and compute the average score $s$ of silhouette coefficients of all clusters. 
As shown in the figure, the score $s$ dramatically improves for all the datasets. 
Especially, the improvements of $s$ in the BC-Alpha and BC-OTC datasets are significant, and the clusters of \method with the loss are more distinguishable. 
All of the results verify the effectiveness of the discriminative loss \lossssl that enhances the disentanglement of \method's node embeddings. 

{\color{blue}
\begin{figure}[!t]
    \centering
    \hspace{5mm}\includegraphics[width=0.3\linewidth]{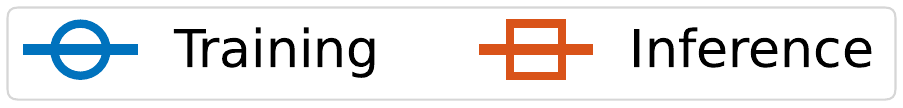}\vspace{-1mm}\\
    \subfigure[Real-world signed graph (Epinions)]{
        \hspace{1mm}
        \includegraphics[width=0.43\textwidth]{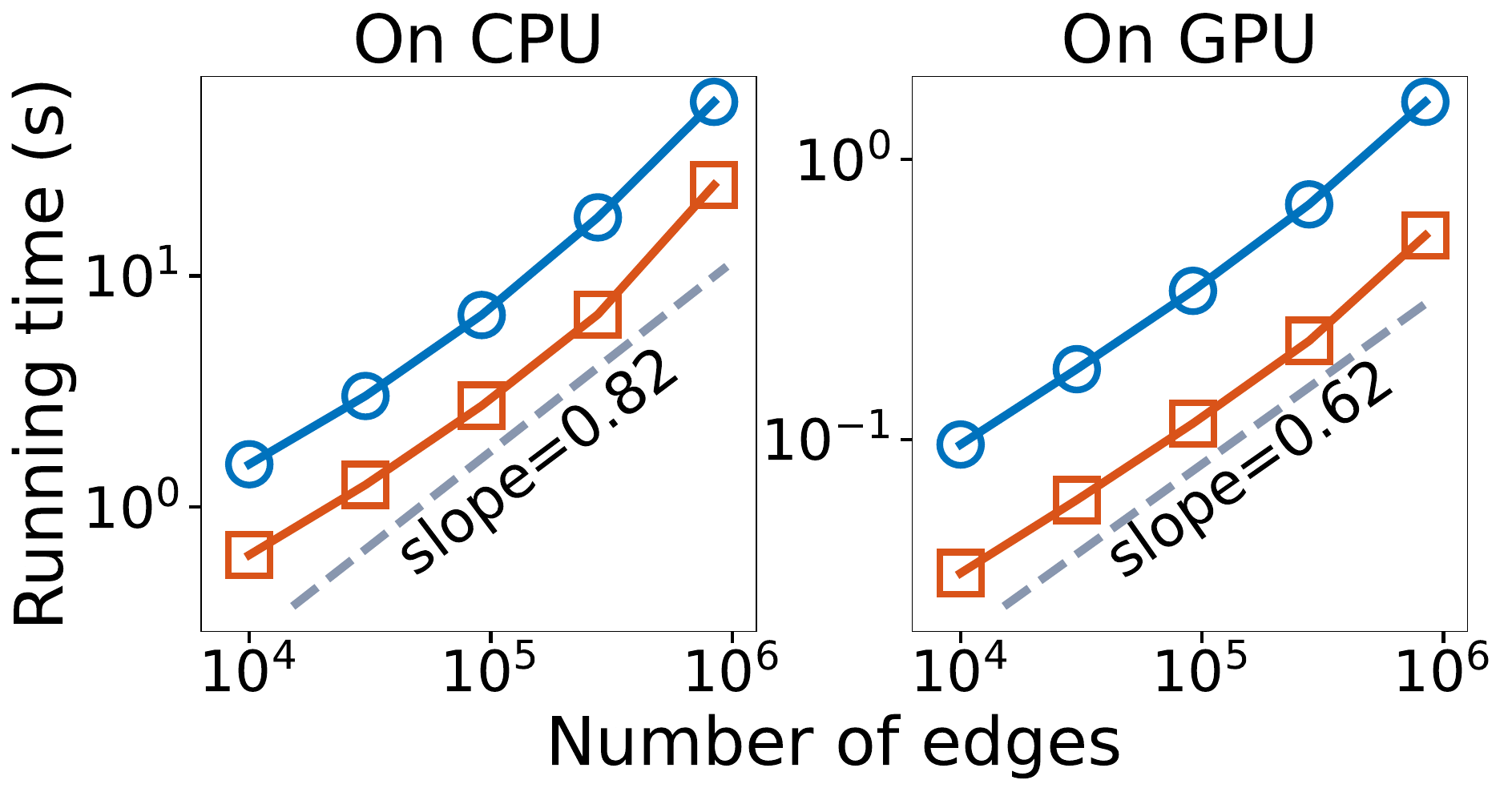}
        \hspace{1mm}
    }
    \subfigure[Synthetic signed graph]{
        \hspace{1mm}
        \includegraphics[width=0.43\textwidth]{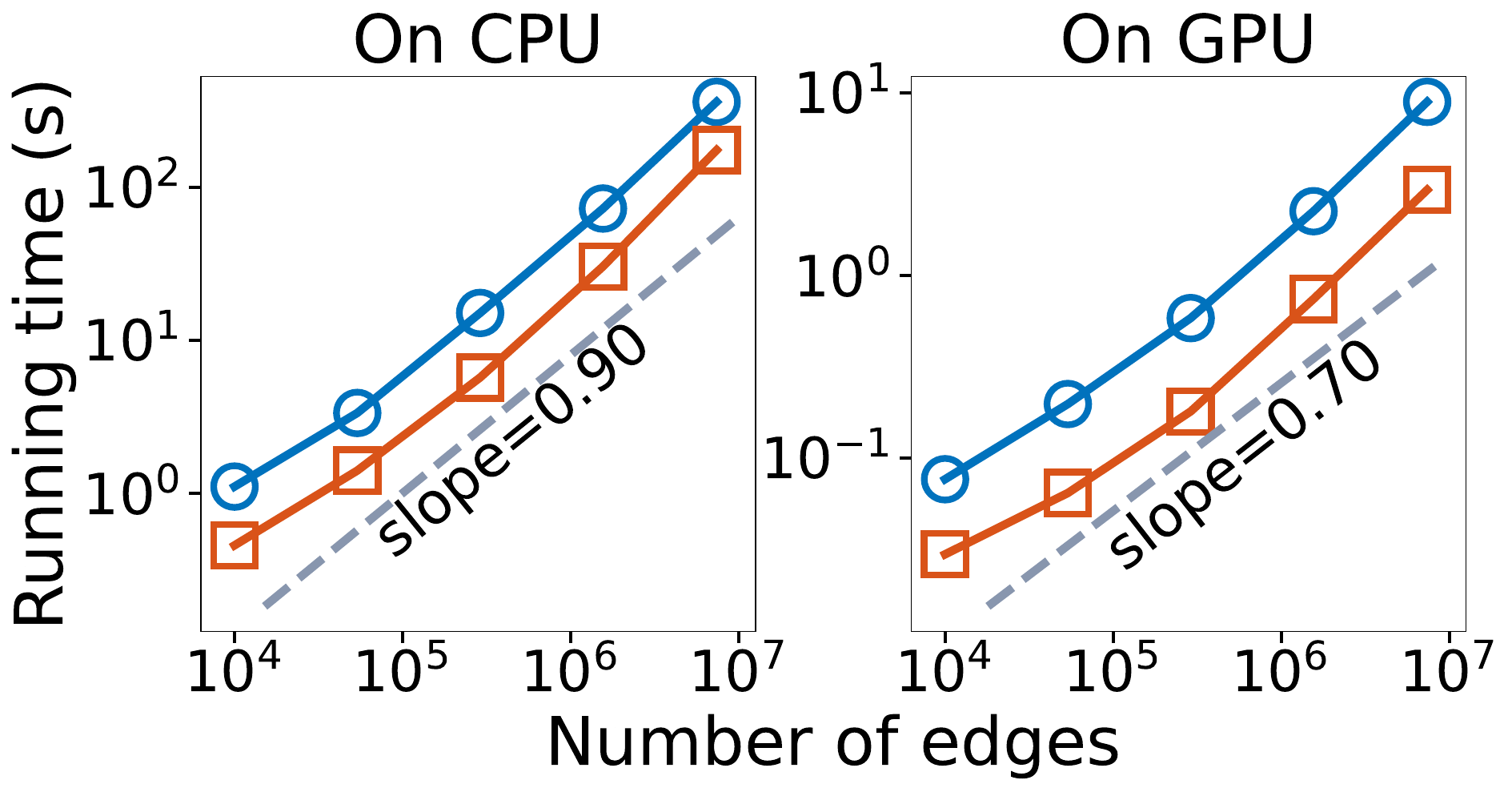}
        \hspace{1mm}
    }
    \caption{
        \label{fig:experiments:scalability}
        Scalability of \method. 
        The training and inference time of \method shows linear scalability w.r.t. the number of edges in both real-world and synthetic signed graphs.
    }
\end{figure}
}

\subsection{Computational Efficiency}
\label{sec:exp:scalability}

In this section, we evaluate the computational efficiency of our proposed \method in terms of scalability and trade-off.

\smallsection{Scalability} We investigate the scalability of \method between its running time and the number $m$ of edges on real-world and synthetic signed graphs.
For the experiment, we select the Epinions dataset which is the largest real-world graph used in this work, and generate a synthetic signed graph whose number of edges is greater about $10 \times$ than that of the Epinions by using a signed graph generator~\cite{Jung2020-tf} to evaluate the scalability on a larger graph.
We then extract principal submatrices by slicing the upper left part of the adjacency matrix so that $m$ ranges from about $10^{4}$ to the original number of edges in each of the real and synthetic graphs, respectively. 
We conduct the experiment on a CPU (i.e., sequential processing) and a GPU (i.e., parallel processing), respectively, to look into the performance in each environment. 
We report the inference or training time per epoch on average (i.e., the average time over 100 epochs).
Note that the training time includes time for the optimization (backward) phase as well as the inference (forward) phase.

As shown in Figure~\ref{fig:experiments:scalability}, the inference time of \method increases linearly with the number $m$ of edges in both real-world and synthetic graphs, which empirically verifies the result of Theorem~\ref{theorem:time}.
Notice that the training time of \method also shows a similar tendency to the inference time. 
Moreover, \method using a GPU is much faster than that using a CPU due to the GPU's parallel processing, showing lower slopes in both settings. 
These results indicate our \method is scalable linearly w.r.t. the number of edges.

\begin{figure}[!t]
    \centering
    \hspace{5mm}\includegraphics[width=\linewidth]{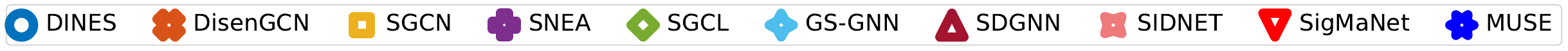}\vspace{-2mm}\\
    \subfigure[BC-Alpha]{
        \hspace{-7mm}
        \includegraphics[width=0.224\linewidth]{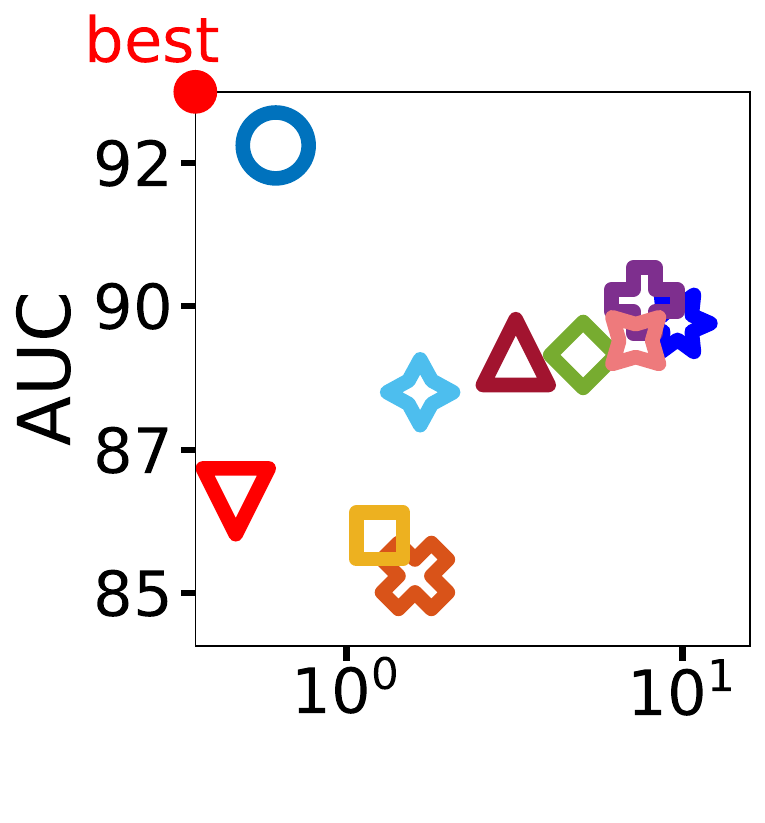}
        \hspace{-2mm}
        \label{fig:experiments:tradeoff:BC-Alpha}
    }
    \subfigure[BC-OTC]{
        \hspace{-2.5mm}
        \includegraphics[width=0.203\linewidth]{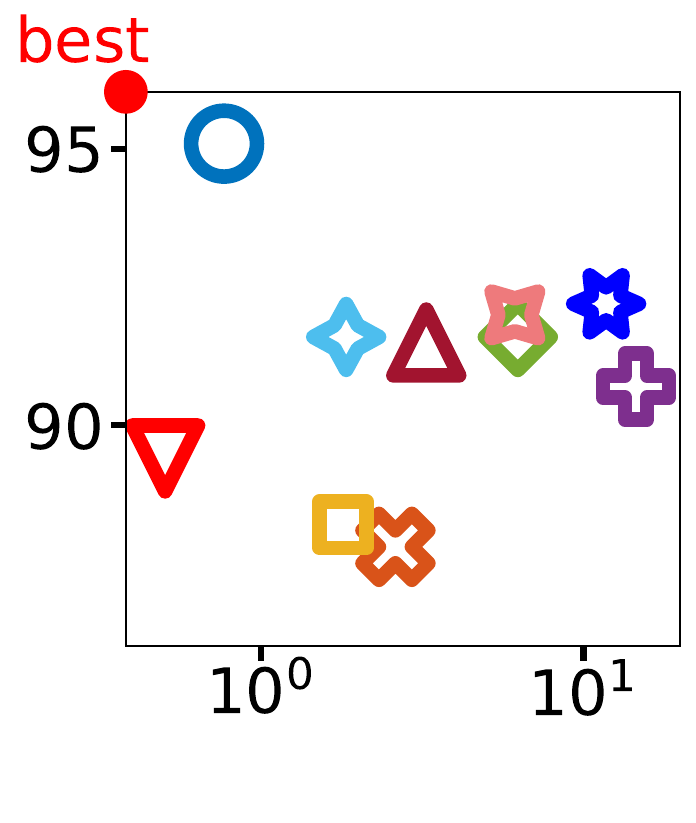}
        \label{fig:experiments:tradeoff:BC-OTC}
    }
    \subfigure[Wiki-RFA]{
        \hspace{-3.5mm}
        \includegraphics[width=0.203\linewidth]{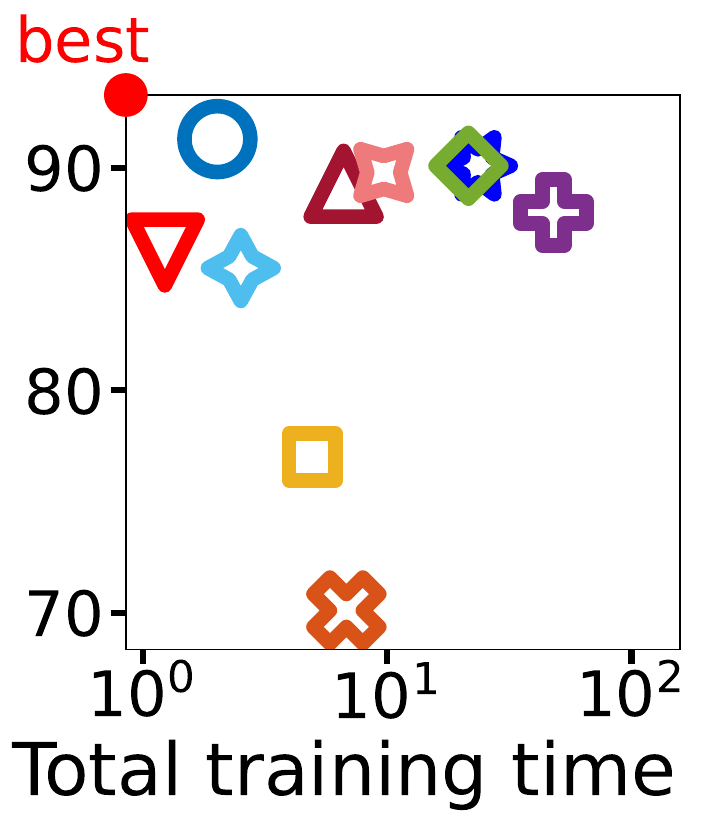}
        \hspace{-3mm}
        \label{fig:experiments:tradeoff:Wiki-Rfa}
    }
    \subfigure[Slashdot]{
        \hspace{-2mm}
        \includegraphics[width=0.203\linewidth]{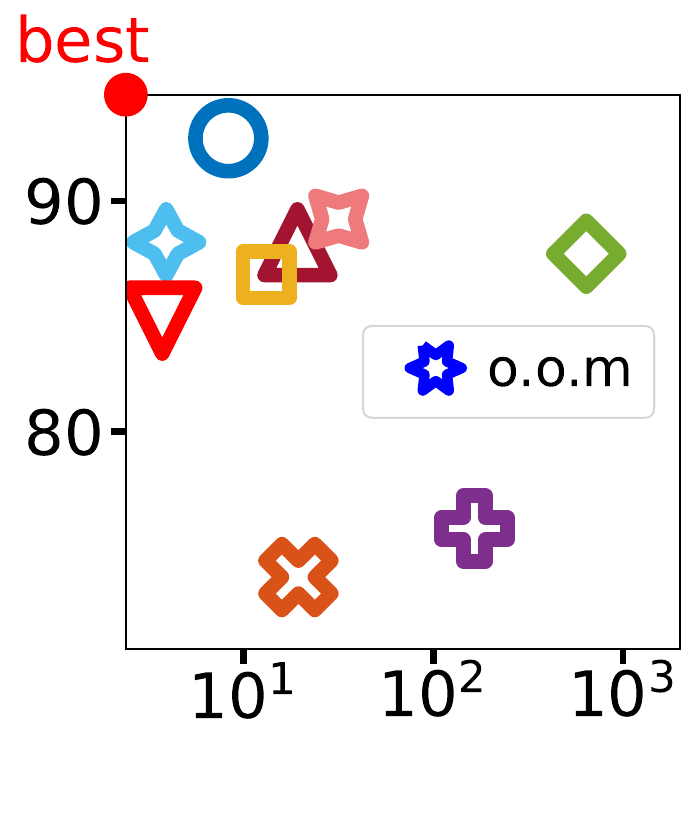}
        \label{fig:experiments:tradeoff:Slashdot}
    }
    \subfigure[Epinions]{
        \hspace{-3mm}
        \includegraphics[width=0.203\linewidth]{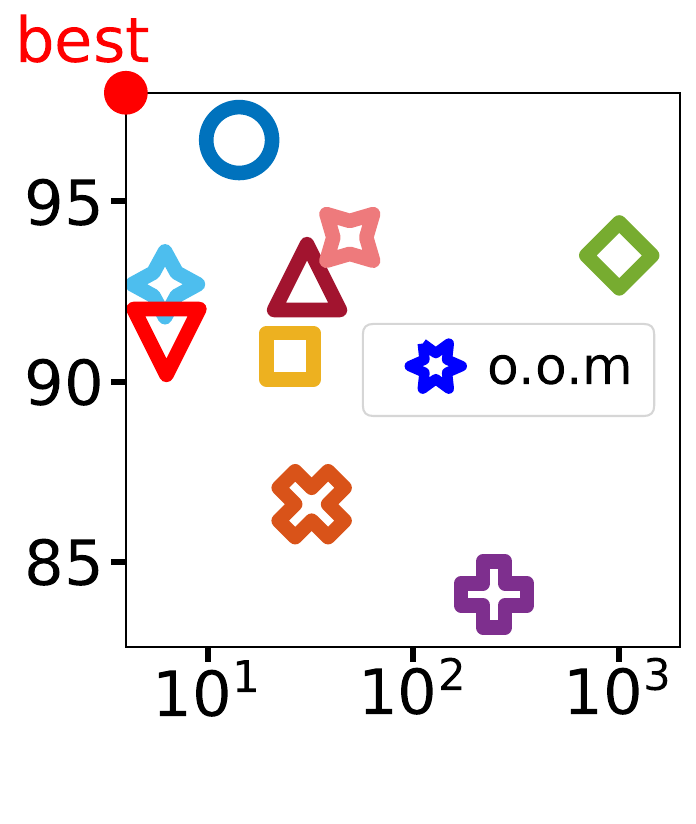}
        \hspace{-5mm}
        \label{fig:experiments:tradeoff:Epinions}
    }
    
    \caption{
        \label{fig:experiments:trade_off}
        Trade-off between accuracy and running time where the upper left marker indicates the best trade-off. 
        Our method \method provides the best accuracy with a fast training time compared to other methods. 
    }
\end{figure}

\smallsection{Accuracy v.s. Training Time} We further analyze the trade-off between accuracy and training time of all GNN methods including \method.  
For each dataset, we train a model for $100$ epochs (i.e., for training, all models are given an equal opportunity to read the dataset in the same number of epochs), and report the total training time on a GPU and the test accuracy in AUC, as shown in Figure~\ref{fig:experiments:trade_off}.
Note \method achieves the best accuracy while showing fast training time compared to other methods (i.e., it is closest to the best point marker across most datasets). 
The reason is that our model leverages the correlation of multiple factors to enhance learning performance, and our approach employs lightweight graph convolutions without requiring the optimization of additional structural losses, in contrast to SGCN, SENA, and SDGNN.

\section{Conclusion}
\label{sec:conclusion}

In this work, we propose \method, a novel method for learning node representations in signed directed graphs without social assumptions. 
Our learning framework utilizes a disentangled architecture, specifically designed to learn diverse latent factors of each node through disentangled signed graph convolution layers.
To capture the signed relationship between two nodes more effectively, we introduce a novel decoder that leverages pairwise correlations among their disentangled factors, and analyze how our decoding strategy is more effective than the traditional one. 
We further adopt a self-supervised factor-wise classification to boost the disentanglement of factors.
Throughout extensive experiments, we demonstrate that \method outperforms other state-of-the-art methods in the link sign prediction task, showing linear scalability and a good trade-off between accuracy and time.

\bibliographystyle{unsrtnat}

\begin{thebibliography}{53}
\providecommand{\natexlab}[1]{#1}
\providecommand{\url}[1]{\texttt{#1}}
\expandafter\ifx\csname urlstyle\endcsname\relax
  \providecommand{\doi}[1]{doi: #1}\else
  \providecommand{\doi}{doi: \begingroup \urlstyle{rm}\Url}\fi

\bibitem[Guha et~al.(2004)Guha, Kumar, Raghavan, and Tomkins]{Guha2004-yz}
Ramanathan~V. Guha, Ravi Kumar, Prabhakar Raghavan, and Andrew Tomkins.
\newblock Propagation of trust and distrust.
\newblock In \emph{Proceedings of the 13th international conference on World
  Wide Web, {WWW} 2004, New York, NY, USA, May 17-20, 2004}, pages 403--412.
  {ACM}, 2004.
\newblock \doi{10.1145/988672.988727}.
\newblock URL \url{https://doi.org/10.1145/988672.988727}.

\bibitem[Kumar et~al.(2016)Kumar, Spezzano, Subrahmanian, and
  Faloutsos]{Kumar2016-lm}
Srijan Kumar, Francesca Spezzano, V.~S. Subrahmanian, and Christos Faloutsos.
\newblock Edge weight prediction in weighted signed networks.
\newblock In \emph{{IEEE} 16th International Conference on Data Mining, {ICDM}
  2016, December 12-15, 2016, Barcelona, Spain}, pages 221--230. {IEEE}
  Computer Society, 2016.
\newblock \doi{10.1109/ICDM.2016.0033}.
\newblock URL \url{https://doi.org/10.1109/ICDM.2016.0033}.

\bibitem[Kunegis et~al.(2009)Kunegis, Lommatzsch, and
  Bauckhage]{Kunegis2009-hw}
J{\'{e}}r{\^{o}}me Kunegis, Andreas Lommatzsch, and Christian Bauckhage.
\newblock The slashdot zoo: mining a social network with negative edges.
\newblock In \emph{Proceedings of the 18th International Conference on World
  Wide Web, {WWW} 2009, Madrid, Spain, April 20-24, 2009}, pages 741--750.
  {ACM}, 2009.
\newblock \doi{10.1145/1526709.1526809}.
\newblock URL \url{https://doi.org/10.1145/1526709.1526809}.

\bibitem[Seo et~al.(2021)Seo, Jeong, Lim, and Shin]{seo2022siren}
Changwon Seo, Kyeong{-}Joong Jeong, Sungsu Lim, and Won{-}Yong Shin.
\newblock Siren: Sign-aware recommendation using graph neural networks.
\newblock \emph{CoRR}, abs/2108.08735, 2021.
\newblock URL \url{https://arxiv.org/abs/2108.08735}.

\bibitem[Leskovec et~al.(2010{\natexlab{a}})Leskovec, Huttenlocher, and
  Kleinberg]{leskovec2010signed}
Jure Leskovec, Daniel~P. Huttenlocher, and Jon~M. Kleinberg.
\newblock Signed networks in social media.
\newblock In \emph{Proceedings of the 28th International Conference on Human
  Factors in Computing Systems, {CHI} 2010, Atlanta, Georgia, USA, April 10-15,
  2010}, pages 1361--1370. {ACM}, 2010{\natexlab{a}}.
\newblock \doi{10.1145/1753326.1753532}.
\newblock URL \url{https://doi.org/10.1145/1753326.1753532}.

\bibitem[Leskovec et~al.(2010{\natexlab{b}})Leskovec, Huttenlocher, and
  Kleinberg]{leskovec2010predicting}
Jure Leskovec, Daniel~P. Huttenlocher, and Jon~M. Kleinberg.
\newblock Predicting positive and negative links in online social networks.
\newblock In \emph{Proceedings of the 19th International Conference on World
  Wide Web, {WWW} 2010, Raleigh, North Carolina, USA, April 26-30, 2010}, pages
  641--650. {ACM}, 2010{\natexlab{b}}.
\newblock \doi{10.1145/1772690.1772756}.
\newblock URL \url{https://doi.org/10.1145/1772690.1772756}.

\bibitem[West et~al.(2014)West, Paskov, Leskovec, and Potts]{West2014-ij}
Robert West, Hristo~S. Paskov, Jure Leskovec, and Christopher Potts.
\newblock Exploiting social network structure for person-to-person sentiment
  analysis.
\newblock \emph{Transactions of the Association for Computational Linguistics},
  2:\penalty0 297--310, 2014.
\newblock \doi{10.1162/tacl\_a\_00184}.
\newblock URL \url{https://doi.org/10.1162/tacl\_a\_00184}.

\bibitem[Tang et~al.(2016{\natexlab{a}})Tang, Chang, Aggarwal, and
  Liu]{tang2016survey}
Jiliang Tang, Yi~Chang, Charu~C. Aggarwal, and Huan Liu.
\newblock A survey of signed network mining in social media.
\newblock \emph{ACM Computing Surveys}, 49\penalty0 (3):\penalty0 42:1--42:37,
  2016{\natexlab{a}}.
\newblock \doi{10.1145/2956185}.
\newblock URL \url{https://doi.org/10.1145/2956185}.

\bibitem[Song et~al.(2015)Song, Meyer, and Tao]{song2015efficient}
Dongjin Song, David~A. Meyer, and Dacheng Tao.
\newblock Efficient latent link recommendation in signed networks.
\newblock In \emph{Proceedings of the 21th {ACM} {SIGKDD} International
  Conference on Knowledge Discovery and Data Mining, Sydney, NSW, Australia,
  August 10-13, 2015}, pages 1105--1114. {ACM}, 2015.
\newblock \doi{10.1145/2783258.2783358}.
\newblock URL \url{https://doi.org/10.1145/2783258.2783358}.

\bibitem[Xu et~al.(2019{\natexlab{a}})Xu, Hu, Wu, and Du]{xu2019link}
Pinghua Xu, Wenbin Hu, Jia Wu, and Bo~Du.
\newblock Link prediction with signed latent factors in signed social networks.
\newblock In \emph{Proceedings of the 25th {ACM} {SIGKDD} International
  Conference on Knowledge Discovery {\&} Data Mining, {KDD} 2019, Anchorage,
  AK, USA, August 4-8, 2019}, pages 1046--1054. {ACM}, 2019{\natexlab{a}}.
\newblock \doi{10.1145/3292500.3330850}.
\newblock URL \url{https://doi.org/10.1145/3292500.3330850}.

\bibitem[Jung et~al.(2020{\natexlab{a}})Jung, Jin, and Kang]{jung2020random}
Jinhong Jung, Woojeong Jin, and U~Kang.
\newblock Random walk-based ranking in signed social networks: model and
  algorithms.
\newblock \emph{Knowledge and Information Systems}, 62\penalty0 (2):\penalty0
  571--610, 2020{\natexlab{a}}.
\newblock \doi{10.1007/s10115-019-01364-z}.
\newblock URL \url{https://doi.org/10.1007/s10115-019-01364-z}.

\bibitem[Li et~al.(2019)Li, Fang, and Zhang]{li2019supervised}
Xiaoming Li, Hui Fang, and Jie Zhang.
\newblock Supervised user ranking in signed social networks.
\newblock In \emph{The Thirty-Third {AAAI} Conference on Artificial
  Intelligence, {AAAI} 2019, The Thirty-First Innovative Applications of
  Artificial Intelligence Conference, {IAAI} 2019, The Ninth {AAAI} Symposium
  on Educational Advances in Artificial Intelligence, {EAAI} 2019, Honolulu,
  Hawaii, USA, January 27 - February 1, 2019}, pages 184--191. {AAAI} Press,
  2019.
\newblock \doi{10.1609/aaai.v33i01.3301184}.
\newblock URL \url{https://doi.org/10.1609/aaai.v33i01.3301184}.

\bibitem[Jung et~al.(2016)Jung, Jin, Sael, and Kang]{jung2016personalized}
Jinhong Jung, Woojeong Jin, Lee Sael, and U~Kang.
\newblock Personalized ranking in signed networks using signed random walk with
  restart.
\newblock In \emph{{IEEE} 16th International Conference on Data Mining, {ICDM}
  2016, December 12-15, 2016, Barcelona, Spain}, pages 973--978. {IEEE}
  Computer Society, 2016.
\newblock \doi{10.1109/ICDM.2016.0122}.
\newblock URL \url{https://doi.org/10.1109/ICDM.2016.0122}.

\bibitem[Tang et~al.(2016{\natexlab{b}})Tang, Aggarwal, and Liu]{tang2016node}
Jiliang Tang, Charu~C. Aggarwal, and Huan Liu.
\newblock Node classification in signed social networks.
\newblock In \emph{Proceedings of the 2016 {SIAM} International Conference on
  Data Mining, Miami, Florida, USA, May 5-7, 2016}, pages 54--62. {SIAM},
  2016{\natexlab{b}}.
\newblock \doi{10.1137/1.9781611974348.7}.
\newblock URL \url{https://doi.org/10.1137/1.9781611974348.7}.

\bibitem[He et~al.(2022)He, Reinert, Wang, and Cucuringu]{he2022sssnet}
Yixuan He, Gesine Reinert, Songchao Wang, and Mihai Cucuringu.
\newblock {SSSNET:} semi-supervised signed network clustering.
\newblock In \emph{Proceedings of the 2022 {SIAM} International Conference on
  Data Mining, {SDM} 2022, Alexandria, VA, USA, April 28-30, 2022}, pages
  244--252. {SIAM}, 2022.
\newblock \doi{10.1137/1.9781611977172.28}.
\newblock URL \url{https://doi.org/10.1137/1.9781611977172.28}.

\bibitem[Kumar et~al.(2014)Kumar, Spezzano, and
  Subrahmanian]{kumar2014accurately}
Srijan Kumar, Francesca Spezzano, and V.~S. Subrahmanian.
\newblock Accurately detecting trolls in slashdot zoo via decluttering.
\newblock In \emph{2014 {IEEE/ACM} International Conference on Advances in
  Social Networks Analysis and Mining, {ASONAM} 2014, Beijing, China, August
  17-20, 2014}, pages 188--195. {IEEE} Computer Society, 2014.
\newblock \doi{10.1109/ASONAM.2014.6921581}.
\newblock URL \url{https://doi.org/10.1109/ASONAM.2014.6921581}.

\bibitem[Yang et~al.(2007)Yang, Cheung, and Liu]{yang2007community}
Bo~Yang, William~K. Cheung, and Jiming Liu.
\newblock Community mining from signed social networks.
\newblock \emph{IEEE transactions on knowledge and data engineering},
  19\penalty0 (10):\penalty0 1333--1348, 2007.
\newblock \doi{10.1109/TKDE.2007.1061}.
\newblock URL \url{https://doi.org/10.1109/TKDE.2007.1061}.

\bibitem[Chu et~al.(2016)Chu, Wang, Pei, Wang, Zhao, and Chen]{chu2016finding}
Lingyang Chu, Zhefeng Wang, Jian Pei, Jiannan Wang, Zijin Zhao, and Enhong
  Chen.
\newblock Finding gangs in war from signed networks.
\newblock In \emph{Proceedings of the 22nd {ACM} {SIGKDD} International
  Conference on Knowledge Discovery and Data Mining, San Francisco, CA, USA,
  August 13-17, 2016}, pages 1505--1514. {ACM}, 2016.
\newblock \doi{10.1145/2939672.2939855}.
\newblock URL \url{https://doi.org/10.1145/2939672.2939855}.

\bibitem[Tzeng et~al.(2020)Tzeng, Ordozgoiti, and Gionis]{tzeng2020discovering}
Ruo{-}Chun Tzeng, Bruno Ordozgoiti, and Aristides Gionis.
\newblock Discovering conflicting groups in signed networks.
\newblock In \emph{Advances in Neural Information Processing Systems 33: Annual
  Conference on Neural Information Processing Systems 2020, NeurIPS 2020,
  December 6-12, 2020, virtual}, 2020.
\newblock URL
  \url{https://proceedings.neurips.cc/paper/2020/hash/7cc538b1337957dae283c30ad46def38-Abstract.html}.

\bibitem[Chen et~al.(2018)Chen, Qian, Liu, and Sun]{ChenQLS18}
Yiqi Chen, Tieyun Qian, Huan Liu, and Ke~Sun.
\newblock "bridge": Enhanced signed directed network embedding.
\newblock In \emph{Proceedings of the 27th {ACM} International Conference on
  Information and Knowledge Management, {CIKM} 2018, Torino, Italy, October
  22-26, 2018}, pages 773--782. {ACM}, 2018.
\newblock \doi{10.1145/3269206.3271738}.
\newblock URL \url{https://doi.org/10.1145/3269206.3271738}.

\bibitem[Yuan et~al.(2017)Yuan, Wu, and Xiang]{yuan2017sne}
Shuhan Yuan, Xintao Wu, and Yang Xiang.
\newblock {SNE:} signed network embedding.
\newblock In \emph{Advances in Knowledge Discovery and Data Mining - 21st
  Pacific-Asia Conference, {PAKDD} 2017, Jeju, South Korea, May 23-26, 2017,
  Proceedings, Part {II}}, volume 10235 of \emph{Lecture Notes in Computer
  Science}, pages 183--195, 2017.
\newblock \doi{10.1007/978-3-319-57529-2\_15}.
\newblock URL \url{https://doi.org/10.1007/978-3-319-57529-2\_15}.

\bibitem[Kim et~al.(2018)Kim, Park, Lee, and Kang]{Kim2018-aa}
Junghwan Kim, Haekyu Park, Ji{-}Eun Lee, and U~Kang.
\newblock {SIDE:} representation learning in signed directed networks.
\newblock In \emph{Proceedings of the 2018 World Wide Web Conference on World
  Wide Web, {WWW} 2018, Lyon, France, April 23-27, 2018}, pages 509--518.
  {ACM}, 2018.
\newblock \doi{10.1145/3178876.3186117}.
\newblock URL \url{https://doi.org/10.1145/3178876.3186117}.

\bibitem[Lee et~al.(2020)Lee, Seo, Han, and Kim]{lee2020asine}
Yeon{-}Chang Lee, Nayoun Seo, Kyungsik Han, and Sang{-}Wook Kim.
\newblock Asine: Adversarial signed network embedding.
\newblock In \emph{Proceedings of the 43rd International {ACM} {SIGIR}
  conference on research and development in Information Retrieval, {SIGIR}
  2020, Virtual Event, China, July 25-30, 2020}, pages 609--618. {ACM}, 2020.
\newblock \doi{10.1145/3397271.3401079}.
\newblock URL \url{https://doi.org/10.1145/3397271.3401079}.

\bibitem[Xu et~al.(2022)Xu, Zhan, Liu, Yu, Du, Wu, and Hu]{xu2022dual}
Pinghua Xu, Yibing Zhan, Liu Liu, Baosheng Yu, Bo~Du, Jia Wu, and Wenbin Hu.
\newblock Dual-branch density ratio estimation for signed network embedding.
\newblock In \emph{{WWW} '22: The {ACM} Web Conference 2022, Virtual Event,
  Lyon, France, April 25 - 29, 2022}, pages 1651--1662. {ACM}, 2022.
\newblock \doi{10.1145/3485447.3512171}.
\newblock URL \url{https://doi.org/10.1145/3485447.3512171}.

\bibitem[Derr et~al.(2018)Derr, Ma, and Tang]{Derr2018-bs}
Tyler Derr, Yao Ma, and Jiliang Tang.
\newblock Signed graph convolutional networks.
\newblock In \emph{{IEEE} International Conference on Data Mining, {ICDM} 2018,
  Singapore, November 17-20, 2018}, pages 929--934. {IEEE} Computer Society,
  2018.
\newblock \doi{10.1109/ICDM.2018.00113}.
\newblock URL \url{https://doi.org/10.1109/ICDM.2018.00113}.

\bibitem[Li et~al.(2020)Li, Tian, Zhang, and Chang]{Li2020-py}
Yu~Li, Yuan Tian, Jiawei Zhang, and Yi~Chang.
\newblock Learning signed network embedding via graph attention.
\newblock In \emph{The Thirty-Fourth {AAAI} Conference on Artificial
  Intelligence, {AAAI} 2020, The Thirty-Second Innovative Applications of
  Artificial Intelligence Conference, {IAAI} 2020, The Tenth {AAAI} Symposium
  on Educational Advances in Artificial Intelligence, {EAAI} 2020, New York,
  NY, USA, February 7-12, 2020}, pages 4772--4779. {AAAI} Press, 2020.
\newblock URL \url{https://ojs.aaai.org/index.php/AAAI/article/view/5911}.

\bibitem[Liu et~al.(2021)Liu, Zhang, Cui, Zhang, Cui, Liu, and Zhu]{Liu2021-np}
Haoxin Liu, Ziwei Zhang, Peng Cui, Yafeng Zhang, Qiang Cui, Jiashuo Liu, and
  Wenwu Zhu.
\newblock Signed graph neural network with latent groups.
\newblock In \emph{{KDD} '21: The 27th {ACM} {SIGKDD} Conference on Knowledge
  Discovery and Data Mining, Virtual Event, Singapore, August 14-18, 2021},
  pages 1066--1075. {ACM}, 2021.
\newblock \doi{10.1145/3447548.3467355}.
\newblock URL \url{https://doi.org/10.1145/3447548.3467355}.

\bibitem[Shu et~al.(2021)Shu, Du, Chang, Chen, Zheng, Xing, and
  Shen]{shu2021sgcl}
Lin Shu, Erxin Du, Yaomin Chang, Chuan Chen, Zibin Zheng, Xingxing Xing, and
  Shaofeng Shen.
\newblock {SGCL:} contrastive representation learning for signed graphs.
\newblock In \emph{{CIKM} '21: The 30th {ACM} International Conference on
  Information and Knowledge Management, Virtual Event, Queensland, Australia,
  November 1 - 5, 2021}, pages 1671--1680. {ACM}, 2021.
\newblock \doi{10.1145/3459637.3482478}.
\newblock URL \url{https://doi.org/10.1145/3459637.3482478}.

\bibitem[Huang et~al.(2021)Huang, Shen, Hou, and Cheng]{Huang2021-wp}
Junjie Huang, Huawei Shen, Liang Hou, and Xueqi Cheng.
\newblock {SDGNN:} learning node representation for signed directed networks.
\newblock In \emph{Thirty-Fifth {AAAI} Conference on Artificial Intelligence,
  {AAAI} 2021, Thirty-Third Conference on Innovative Applications of Artificial
  Intelligence, {IAAI} 2021, The Eleventh Symposium on Educational Advances in
  Artificial Intelligence, {EAAI} 2021, Virtual Event, February 2-9, 2021},
  pages 196--203. {AAAI} Press, 2021.
\newblock URL \url{https://ojs.aaai.org/index.php/AAAI/article/view/16093}.

\bibitem[Jung et~al.(2022)Jung, Yoo, and Kang]{Jung2022-pm}
Jinhong Jung, Jaemin Yoo, and U.~Kang.
\newblock Signed random walk diffusion for effective representation learning in
  signed graphs.
\newblock \emph{PLOS ONE}, 17\penalty0 (3):\penalty0 1--19, 03 2022.
\newblock \doi{10.1371/journal.pone.0265001}.
\newblock URL \url{https://doi.org/10.1371/journal.pone.0265001}.

\bibitem[Yan et~al.(2023)Yan, Zhang, Xie, Jin, and Zhang]{Yan2023-ta}
Dengcheng Yan, Youwen Zhang, Wenxin Xie, Ying Jin, and Yiwen Zhang.
\newblock {MUSE:} multi-faceted attention for signed network embedding.
\newblock \emph{Neurocomputing}, 519:\penalty0 36--43, 2023.
\newblock \doi{10.1016/j.neucom.2022.11.021}.
\newblock URL \url{https://doi.org/10.1016/j.neucom.2022.11.021}.

\bibitem[Fiorini et~al.(2022)Fiorini, Coniglio, Ciavotta, and
  Messina]{Fiorini2022-ef}
Stefano Fiorini, Stefano Coniglio, Michele Ciavotta, and Enza Messina.
\newblock Sigmanet: One laplacian to rule them all.
\newblock \emph{CoRR}, abs/2205.13459, 2022.
\newblock \doi{10.48550/arXiv.2205.13459}.
\newblock URL \url{https://doi.org/10.48550/arXiv.2205.13459}.

\bibitem[Cartwright and Harary(1956)]{cartwright1956structural}
Dorwin Cartwright and Frank Harary.
\newblock Structural balance: a generalization of heider's theory.
\newblock \emph{Psychological review}, 63\penalty0 (5):\penalty0 277, 1956.

\bibitem[Falher(2018)]{le2018characterizing}
G{\'{e}}raud~Le Falher.
\newblock \emph{Characterizing Edges in Signed and Vector-Valued Graphs.
  (Caract{\'{e}}risation des ar{\^{e}}tes dans les graphes sign{\'{e}}s et
  attribu{\'{e}}s)}.
\newblock PhD thesis, Lille University of Science and Technology, France, 2018.
\newblock URL \url{https://tel.archives-ouvertes.fr/tel-01824215}.

\bibitem[Davis(1967)]{davis1967clustering}
James~A Davis.
\newblock Clustering and structural balance in graphs.
\newblock \emph{Human relations}, 20\penalty0 (2):\penalty0 181--187, 1967.

\bibitem[Ma et~al.(2019)Ma, Cui, Kuang, Wang, and Zhu]{Ma2019-ce}
Jianxin Ma, Peng Cui, Kun Kuang, Xin Wang, and Wenwu Zhu.
\newblock Disentangled graph convolutional networks.
\newblock In \emph{Proceedings of the 36th International Conference on Machine
  Learning, {ICML} 2019, 9-15 June 2019, Long Beach, California, {USA}},
  volume~97 of \emph{Proceedings of Machine Learning Research}, pages
  4212--4221. {PMLR}, 2019.
\newblock URL \url{http://proceedings.mlr.press/v97/ma19a.html}.

\bibitem[Liu et~al.(2020)Liu, Wang, Wu, and Xiao]{Liu2020-yd}
Yanbei Liu, Xiao Wang, Shu Wu, and Zhitao Xiao.
\newblock Independence promoted graph disentangled networks.
\newblock In \emph{The Thirty-Fourth {AAAI} Conference on Artificial
  Intelligence, {AAAI} 2020, The Thirty-Second Innovative Applications of
  Artificial Intelligence Conference, {IAAI} 2020, The Tenth {AAAI} Symposium
  on Educational Advances in Artificial Intelligence, {EAAI} 2020, New York,
  NY, USA, February 7-12, 2020}, pages 4916--4923. {AAAI} Press, 2020.
\newblock URL \url{https://ojs.aaai.org/index.php/AAAI/article/view/5929}.

\bibitem[Kipf and Welling(2017)]{KipfW17}
Thomas~N. Kipf and Max Welling.
\newblock Semi-supervised classification with graph convolutional networks.
\newblock In \emph{5th International Conference on Learning Representations,
  {ICLR} 2017, Toulon, France, April 24-26, 2017, Conference Track
  Proceedings}. OpenReview.net, 2017.
\newblock URL \url{https://openreview.net/forum?id=SJU4ayYgl}.

\bibitem[Klicpera et~al.(2019)Klicpera, Bojchevski, and
  G{\"{u}}nnemann]{gasteiger2018predict}
Johannes Klicpera, Aleksandar Bojchevski, and Stephan G{\"{u}}nnemann.
\newblock Predict then propagate: Graph neural networks meet personalized
  pagerank.
\newblock In \emph{7th International Conference on Learning Representations,
  {ICLR} 2019, New Orleans, LA, USA, May 6-9, 2019}. OpenReview.net, 2019.
\newblock URL \url{https://openreview.net/forum?id=H1gL-2A9Ym}.

\bibitem[Wang et~al.(2017)Wang, Tang, Aggarwal, Chang, and Liu]{Wang2017-jw}
Suhang Wang, Jiliang Tang, Charu~C. Aggarwal, Yi~Chang, and Huan Liu.
\newblock Signed network embedding in social media.
\newblock In \emph{Proceedings of the 2017 {SIAM} International Conference on
  Data Mining, Houston, Texas, USA, April 27-29, 2017}, pages 327--335. {SIAM},
  2017.
\newblock \doi{10.1137/1.9781611974973.37}.
\newblock URL \url{https://doi.org/10.1137/1.9781611974973.37}.

\bibitem[Wang et~al.(2020)Wang, Jin, Zhang, He, Xu, and Chua]{Wang2020-jf}
Xiang Wang, Hongye Jin, An~Zhang, Xiangnan He, Tong Xu, and Tat{-}Seng Chua.
\newblock Disentangled graph collaborative filtering.
\newblock In \emph{Proceedings of the 43rd International {ACM} {SIGIR}
  conference on research and development in Information Retrieval, {SIGIR}
  2020, Virtual Event, China, July 25-30, 2020}, pages 1001--1010. {ACM}, 2020.
\newblock \doi{10.1145/3397271.3401137}.
\newblock URL \url{https://doi.org/10.1145/3397271.3401137}.

\bibitem[Zhao et~al.(2022)Zhao, Zhang, and Wang]{zhao2022exploring}
Tianxiang Zhao, Xiang Zhang, and Suhang Wang.
\newblock Exploring edge disentanglement for node classification.
\newblock In \emph{{WWW} '22: The {ACM} Web Conference 2022, Virtual Event,
  Lyon, France, April 25 - 29, 2022}, pages 1028--1036. {ACM}, 2022.
\newblock \doi{10.1145/3485447.3511929}.
\newblock URL \url{https://doi.org/10.1145/3485447.3511929}.

\bibitem[Hamilton et~al.(2017)Hamilton, Ying, and Leskovec]{Hamilton2017-bh}
William~L. Hamilton, Zhitao Ying, and Jure Leskovec.
\newblock Inductive representation learning on large graphs.
\newblock In \emph{Advances in Neural Information Processing Systems 30: Annual
  Conference on Neural Information Processing Systems 2017, December 4-9, 2017,
  Long Beach, CA, {USA}}, pages 1024--1034, 2017.
\newblock URL
  \url{https://proceedings.neurips.cc/paper/2017/hash/5dd9db5e033da9c6fb5ba83c7a7ebea9-Abstract.html}.

\bibitem[Xu et~al.(2019{\natexlab{b}})Xu, Hu, Leskovec, and Jegelka]{Xu2018-sy}
Keyulu Xu, Weihua Hu, Jure Leskovec, and Stefanie Jegelka.
\newblock How powerful are graph neural networks?
\newblock In \emph{7th International Conference on Learning Representations,
  {ICLR} 2019, New Orleans, LA, USA, May 6-9, 2019}. OpenReview.net,
  2019{\natexlab{b}}.
\newblock URL \url{https://openreview.net/forum?id=ryGs6iA5Km}.

\bibitem[Perozzi et~al.(2014)Perozzi, Al{-}Rfou, and Skiena]{Perozzi2014-qc}
Bryan Perozzi, Rami Al{-}Rfou, and Steven Skiena.
\newblock Deepwalk: online learning of social representations.
\newblock In \emph{The 20th {ACM} {SIGKDD} International Conference on
  Knowledge Discovery and Data Mining, {KDD} '14, New York, NY, {USA} - August
  24 - 27, 2014}, pages 701--710. {ACM}, 2014.
\newblock \doi{10.1145/2623330.2623732}.
\newblock URL \url{https://doi.org/10.1145/2623330.2623732}.

\bibitem[Tang et~al.(2015)Tang, Qu, Wang, Zhang, Yan, and Mei]{Tang2015-xs}
Jian Tang, Meng Qu, Mingzhe Wang, Ming Zhang, Jun Yan, and Qiaozhu Mei.
\newblock {LINE:} large-scale information network embedding.
\newblock In \emph{Proceedings of the 24th International Conference on World
  Wide Web, {WWW} 2015, Florence, Italy, May 18-22, 2015}, pages 1067--1077.
  {ACM}, 2015.
\newblock \doi{10.1145/2736277.2741093}.
\newblock URL \url{https://doi.org/10.1145/2736277.2741093}.

\bibitem[Grover and Leskovec(2016)]{Grover2016-qz}
Aditya Grover and Jure Leskovec.
\newblock node2vec: Scalable feature learning for networks.
\newblock In \emph{Proceedings of the 22nd {ACM} {SIGKDD} International
  Conference on Knowledge Discovery and Data Mining, San Francisco, CA, USA,
  August 13-17, 2016}, pages 855--864. {ACM}, 2016.
\newblock \doi{10.1145/2939672.2939754}.
\newblock URL \url{https://doi.org/10.1145/2939672.2939754}.

\bibitem[Bradley(1997)]{Bradley1997-oi}
Andrew~P. Bradley.
\newblock The use of the area under the {ROC} curve in the evaluation of
  machine learning algorithms.
\newblock \emph{Pattern Recognition}, 30\penalty0 (7):\penalty0 1145--1159,
  1997.
\newblock \doi{10.1016/S0031-3203(96)00142-2}.
\newblock URL \url{https://doi.org/10.1016/S0031-3203(96)00142-2}.

\bibitem[Taha and Hanbury(2015)]{Taha2015-oi}
Abdel~Aziz Taha and Allan Hanbury.
\newblock Metrics for evaluating {3D} medical image segmentation: analysis,
  selection, and tool.
\newblock \emph{BMC medical imaging}, 15:\penalty0 29, 2015.

\bibitem[Leskovec et~al.(2010{\natexlab{c}})Leskovec, Huttenlocher, and
  Kleinberg]{Leskovec2010-vk}
Jure Leskovec, Daniel~P. Huttenlocher, and Jon~M. Kleinberg.
\newblock Predicting positive and negative links in online social networks.
\newblock In \emph{Proceedings of the 19th International Conference on World
  Wide Web, {WWW} 2010, Raleigh, North Carolina, USA, April 26-30, 2010}, pages
  641--650. {ACM}, 2010{\natexlab{c}}.
\newblock \doi{10.1145/1772690.1772756}.
\newblock URL \url{https://doi.org/10.1145/1772690.1772756}.

\bibitem[Van~der Maaten and Hinton(2008)]{van2008visualizing}
Laurens Van~der Maaten and Geoffrey Hinton.
\newblock Visualizing data using {t-SNE}.
\newblock \emph{Journal of machine learning research}, 9\penalty0 (11), 2008.

\bibitem[Rousseeuw(1987)]{Rousseeuw1987-qe}
Peter~J Rousseeuw.
\newblock Silhouettes: A graphical aid to the interpretation and validation of
  cluster analysis.
\newblock \emph{Journal of computational and applied mathematics}, 20:\penalty0
  53--65, 1987.

\bibitem[Jung et~al.(2020{\natexlab{b}})Jung, Park, and Kang]{Jung2020-tf}
Jinhong Jung, Ha{-}Myung Park, and U~Kang.
\newblock Balansing: Fast and scalable generation of realistic signed networks.
\newblock In \emph{Proceedings of the 23rd International Conference on
  Extending Database Technology, {EDBT} 2020, Copenhagen, Denmark, March 30 -
  April 02, 2020}, pages 193--204. OpenProceedings.org, 2020{\natexlab{b}}.
\newblock \doi{10.5441/002/edbt.2020.18}.
\newblock URL \url{https://doi.org/10.5441/002/edbt.2020.18}.

\end{thebibliography}

\newpage
\appendix
\section{Appendix}
\subsection{Time Complexity Analysis}
\label{appendix:analysis:time}

\begin{table}[!ht]
\begin{center}
\begin{minipage}{\textwidth}
\small
\caption{
Time complexity of each component of \method where $L$, $d$, and $K$ are hyperparameters (see Table~\ref{tab:symbols}), and they are set to constants much smaller than the numbers $n$ and $m$ of nodes and edges, respectively.
}
\label{tab:time_complexities}
\begin{tabular*}{\textwidth}{@{\extracolsep{\fill}}ccc@{\extracolsep{\fill}}}
    \hline\toprule
\textbf{Components} & \textbf{References} & \textbf{Time Complexities} \\
\midrule
Encoder & Algorithm~\ref{alg:encoder} & $O(L\frac{d^2}{K}m + L\frac{d^2}{K}n)$ \\
Decoder & Algorithm~\ref{alg:decoder} & $O(dKm)$ \\
$\loss_{\texttt{bce}}$ & Equation~\eqref{eq:bce} & $O(m)$ \\
\lossssl & Equation~\eqref{eq:disc} & $O(dKn)$ \\
\midrule
\multicolumn{2}{c}{\textbf{Total Complexity}} & $O(L\frac{d^2}{K}m + L\frac{d^2}{K}n + dKn)$ \\
\multicolumn{2}{c}{\textbf{Simplified Complexity}} & $O(m + n)$\\
\bottomrule\hline
\end{tabular*}
\end{minipage}
\end{center}
\end{table}

\begin{lemma}[Time Complexity of Encoder of \method]
	\label{lemma:encoder}
	The time complexity of Algorithm~\ref{alg:encoder} takes $O(L\frac{d^2}{K}m + L\frac{d^2}{K}n)$ time. 
\begin{proof*}
	For each node $u$, the initial disentanglement (lines~\ref{alg:encoder:initialize:start}-\ref{alg:encoder:initialize:end}) takes $O(d^2\card{\neigh_{u}})$ time (i.e., $O(d^2m)$ time is required for all nodes).
    For each factor $k$, the \ourconv layer consists of three operations called $\texttt{aggregate}_k$, $\texttt{update}_k$, and $\texttt{normalize}_k$.
    For the $\texttt{aggregate}_k(\cdot)$ (line~\ref{alg:encoder:dsgconv:message}), the sum, mean, and max functions take $O(\card{\neigh_{u}}\frac{d}{K})$ time while the attention function requires $O(\card{\neigh_{u}}\frac{d^2}{K^2})$ time where $\card{\neigh_{u}^{\delta}} \leq \card{\neigh_{u}}$.
    Considering the aforementioned aggregators, the time complexity of $\texttt{aggregate}_k(\cdot)$ is up to $O(\card{\neigh_{u}}\frac{d^2}{K^2})$. 
    The $\texttt{update}_k(\cdot)$ and $\texttt{normalize}_k(\cdot)$ take $O(\frac{d^2}{K^2})$ and $O(\frac{d}{K})$ time due to the fully-connected layer and the $l_2$-normalization, respectively.
    Overall, the encoder takes $O(L\frac{d^2}{K}\card{\neigh_{u}}+L\frac{d^2}{K})$ time for each node as it repeats the \texttt{dsg-conv} layer $LK$ times. 
	For all nodes, it exhibits $O(L\frac{d^2}{K}m + L\frac{d^2}{K}n)$ time cost because ${\textstyle\sum}_{u\in\vertex}\card{\neigh_{u}} = O(m)$.
\end{proof*}
\end{lemma}

\begin{lemma}[Time Complexity of Decoder of \method]
	\label{lemma:decoder}
	The time complexity of Algorithm~\ref{alg:decoder} takes $O(dKm)$ time. 
\begin{proof*}
	For each edge $u \rightarrow v$, it takes $O(d)$ time to build $\mat{Z}_{u}$ and $\mat{Z}_{v}$ (line~\ref{alg:decoder:concat}).
	Computing the correlations takes $O(dK)$ time where each inner product takes $O(\frac{d}{K})$ time (line~\ref{alg:decoder:corr}). 
	It consumes $O(K^2)$ time to compute the probability $p_{uv}$ (line~\ref{alg:decoder:prob}). 
	For all given edges (lines~\ref{alg:decoder:for:start}-\ref{alg:decoder:for:end}), our decoder takes $O(dKm + K^2m) = O(dKm)$ time where $K \leq d$.   
\end{proof*}
\end{lemma}


\subsection{Details of Datasets}
\label{appendix:exp:data}

We describe more details of the datasets used in the paper as follows:
\begin{itemize}[leftmargin=*]
    \item {
        \textbf{BC-Alpha}\footnote{\url{https://snap.stanford.edu/data/soc-sign-bitcoin-alpha.html}} and \textbf{BC-OTC}\footnote{\url{https://snap.stanford.edu/data/soc-sign-bitcoin-otc.html}}~\cite{Kumar2016-lm}: 
        These are who-trusts-whom networks of users on Bitcoin platforms where transactions occur anonymously. 
        They allow users to give other users a score between $-10$ (distrust) and $10$ (trust) to reveal fraudulent users.
        As in~\cite{Huang2021-wp,Jung2022-pm}, we also regard links whose scores are higher than zero as positive edges; otherwise, they are considered negative edges. 
    }
    \item {
        \textbf{Wiki-RFA}\footnote{\url{https://snap.stanford.edu/data/wiki-RfA.html}}~\cite{West2014-ij}: This is a signed network in Wikipedia where users vote for administrator candidates.
        The users can cast for or against candidates. 
        The site allowed users to express neutral opinions about the candidates, but few users gave neutral votes. 
        As in~\cite{ChenQLS18,Huang2021-wp}, we also eliminate the neutral votes.
    }
    \item {
        \textbf{Slashdot}\footnote{\url{http://konect.cc/networks/slashdot-zoo}}~\cite{Kunegis2009-hw}: This is a signed social network on Slashdot, a technology news website
        The site introduced a tagging system that allows users to tag other users as friends or foes; thus, the signed graph is represented as trust relationships between the users. 
    }
    \item {
        \textbf{Epinions}\footnote{\url{https://snap.stanford.edu/data/soc-sign-epinions.html}}~\cite{Guha2004-yz}: This is an online social network on Epinions, a general consumer review site. 
        On the website, users can decide whether to trust each other, and the signed graph represents such trust relationships. 
    }
\end{itemize}

Note that following previous work on signed GNNs~\cite{Li2020-py,Liu2021-np,Jung2022-pm}, we utilize truncated singular vector decomposition (TSVD) to generate node features because the aforementioned datasets do not contain initial node features.
Specifically, suppose $\mat{X}\in \mathbb{R}^{n \times d}$ denotes a matrix of node features. 
Then, we set $\mat{X} = \mat{U} \mat{\Sigma}_d$ where a signed adjacency matrix $\mat{A}$ is decomposed by TSVD as $\mat{A} \simeq \mat{U} \mat{\Sigma}_{d} \matt{V}$. 
We follow their setup and extract the initial node features with target rank $d = 64$ for each dataset.

\subsection{Hyperparameters of \method}
\label{appendix:hyperparam}

We report the results of the cross-validation that searches for the hyperparameters of \method (see the symbols in Table~\ref{tab:symbols}).
After repeating the experiments with $10$ different random seeds, we select one validated setting that produces a test accuracy closest to the average test accuracy for each metric. 
The searched hyperparameters are summarized in Tables~\ref{tab:hyperparam:auc}~and~\ref{tab:hyperparam:f1} for AUC and Macro-F1, respectively.

\def\arraystretch{1.1} 
\setlength{\tabcolsep}{2pt} 
\begin{table}[!ht]
\begin{center}
\begin{minipage}{\textwidth}
\small
\caption{
Searched hyperparameters of \method in terms of AUC.  
}
\label{tab:hyperparam:auc}
\begin{tabular*}{\textwidth}{@{\extracolsep{\fill}}ccccccc}
\hline\toprule 
\textbf{Dataset} & $L$ & $K$ & $\dout$ & \lambdassl & $\eta$ & \lambdareg \\
\midrule
BC-Alpha & 2 & 8 & 64 & 0.1 & 0.01 & 0.005 \\
BC-OTC & 2 & 8 & 64 & 0.1 & 0.005 & 0.01 \\
Wiki-RFA & 2 & 8 & 64 & 0.1 & 0.005 & 0.005 \\
Slashdot & 2 & 8 & 64 & 0.5 & 0.01 & 0.01 \\
Epinions & 2 & 8 & 64 & 0.1 & 0.005 & 0.005\\
\bottomrule\hline
\end{tabular*}
\end{minipage}
\end{center}
\end{table}

\vspace{-6mm}
\def\arraystretch{1.1} 
\setlength{\tabcolsep}{2pt} 
\begin{table}[!ht]
\begin{center}
\begin{minipage}{\textwidth}
\small
\caption{
Searched hyperparameters of \method in terms of Macro-F1.  
}
\label{tab:hyperparam:f1}
\begin{tabular*}{\textwidth}{@{\extracolsep{\fill}}ccccccc}
\hline\toprule 
\textbf{Dataset} & $L$ & $K$ & $\dout$ & \lambdassl & $\eta$ & \lambdareg \\
\midrule
BC-Alpha & 2 & 16 & 64 & 0.1 & 0.005 & 0.005 \\
BC-OTC & 2 & 8 & 64 & 0.1 & 0.005 & 0.005 \\
Wiki-RFA & 2 & 16 & 64 & 0.1 & 0.005 & 0.005 \\
Slashdot & 2 & 8 & 64 & 0.1 & 0.005 & 0.01 \\
Epinions & 2 & 8 & 64 & 0.1 & 0.005 & 0.01\\
\bottomrule\hline
\end{tabular*}
\end{minipage}
\end{center}
\end{table}


\subsection{Implementation Information of Competitors}
\label{appendix:url}
For our experiments, we utilized the open-source implementation provided by the authors for the competing methods, and the URLs for each implementation are as follows:
{
\begin{itemize}[leftmargin=*]
	\item \textbf{SNE}: \url{https://bit.ly/42gn9S0}
	\item \textbf{SIDE}: \url{https://datalab.snu.ac.kr/side/resources/side.zip}
	\item \textbf{BESIDE}: \url{https://github.com/yqc01/BESIDE}
	\item \textbf{SLF}: \url{https://github.com/WHU-SNA/SLF}
	\item \textbf{ASiNE}: \url{https://github.com/yeonchang/ASiNE}
	\item \textbf{DDRE}: \url{https://github.com/XphYYYYYYY/DDRE}
	\item \textbf{SGCN}: \url{https://github.com/benedekrozemberczki/SGCN}
	\item \textbf{DisenGCN}: \url{https://github.com/snudatalab/SidNet}
	\item \textbf{SNEA}: \url{https://github.com/liyu1990/snea}
	\item \textbf{SGCL}: \url{https://github.com/xi0927/SGCL}
	\item \textbf{GS-GNN}: \url{https://github.com/haoxin1998/GS-GNN}
	\item \textbf{SDGNN}: \url{https://bit.ly/45Ifmzj}
	\item \textbf{SIDNET}: \url{https://github.com/snudatalab/SidNet}
    \item \textbf{MUSE\footnote{The original code for MUSE was not publicly available; thus, we carefully implemented MUSE based on the description in~\cite{Yan2023-ta}, and made our implementation open-source for reproducibility.}}: \url{https://bit.ly/45TLKPv}
	\item \textbf{SigMaNet}: \url{https://github.com/Stefa1994/SigMaNet}
\end{itemize}
}

\subsection{Properties of Learning Models}
\label{appendix:properties}

We compare \method with other learning models in terms of the following properties:
\begin{enumerate}[leftmargin=13mm, label={\bf(P\arabic*)}]
    \setlength\itemsep{0.1em}
    \item{Does a model use the information from edge signs for its learning? \label{prop:1}}
    \item{Does a model consider the direction of an edge for its learning?\label{prop:2}}
    \item{Does a model train a signed graph and the link sign prediction task in an end-to-end manner?\label{prop:3}}
    \item{Is a model free from social theories (e.g., the balance or status theory) for designing its input, model, convolution, or loss?\label{prop:4}}
    \item{Does a model produce disentangled node representations?\label{prop:5}}
\end{enumerate}

We summarize the comparisons in Table~\ref{tab:properties}. Note that only our method \method satisfies all of the properties. 

\setlength{\tabcolsep}{4.7pt}
\vspace{-2mm}
\begin{table}[!h]
	\centering
	\small
	\caption{\centering Properties of learning models related to \method.}
	\label{tab:properties}
	\begin{tabular}{ccccccc}
\hline\toprule
\textbf{Category} & \textbf{Methods} & \ref{prop:1} & \ref{prop:2} & \ref{prop:3} & \ref{prop:4} & \ref{prop:5} \\
\midrule 
\multirow{6}{*}{\begin{tabular}[c]{@{}c@{}}Network \\ Embedding\end{tabular}} & 
SNE~\cite{yuan2017sne} & \cmark & \cmark & \xmark & \cmark & \xmark \\
 & SIDE~\cite{Kim2018-aa} & \cmark & \cmark & \xmark & \xmark & \xmark \\
 & BESIDE~\cite{ChenQLS18} & \cmark & \cmark & \xmark & \xmark & \xmark \\
 & SLF~\cite{xu2019link} & \cmark & \cmark & \xmark & \cmark & \xmark \\
& ASiNE~\cite{lee2020asine} & \cmark & \xmark & \xmark & \xmark & \xmark \\
 & DDRE~\cite{xu2022dual} & \cmark & \cmark & \xmark & \xmark & \xmark \\
\midrule
\multirow{8}{*}{\begin{tabular}[c]{@{}c@{}}Graph \\ Neural \\ Networks\end{tabular}} & SGCN~\cite{Derr2018-bs} & \cmark & \xmark & \cmark & \xmark & \xmark \\
 & DisenGCN~\cite{Ma2019-ce} & \xmark & \xmark & \cmark & \cmark & \cmark \\
 & SNEA~\cite{Li2020-py} & \cmark & \xmark & \cmark & \xmark & \xmark \\
 & GS-GNN~\cite{Liu2021-np} & \cmark & \xmark & \cmark & \cmark & \xmark \\
 & SGCL~\cite{shu2021sgcl} & \cmark & \cmark & \cmark & \xmark & \xmark \\
 & SDGNN~\cite{Huang2021-wp} & \cmark & \cmark & \cmark & \xmark & \xmark \\
 & SIDNET~\cite{Jung2022-pm} & \cmark & \cmark & \cmark & \xmark & \xmark \\
 & MUSE~\cite{Yan2023-ta} & \cmark & \xmark & \cmark & \xmark & \cmark \\
 & SigMaNet~\cite{Fiorini2022-ef} & \cmark & \cmark & \cmark & \cmark & \xmark \\
 & \textbf{DINES (ours)} & \cmark & \cmark & \cmark & \cmark & \cmark \\
\bottomrule\hline
	\end{tabular}
\end{table}

\begin{figure}[!b]
    \centering
    \includegraphics[width=\linewidth]{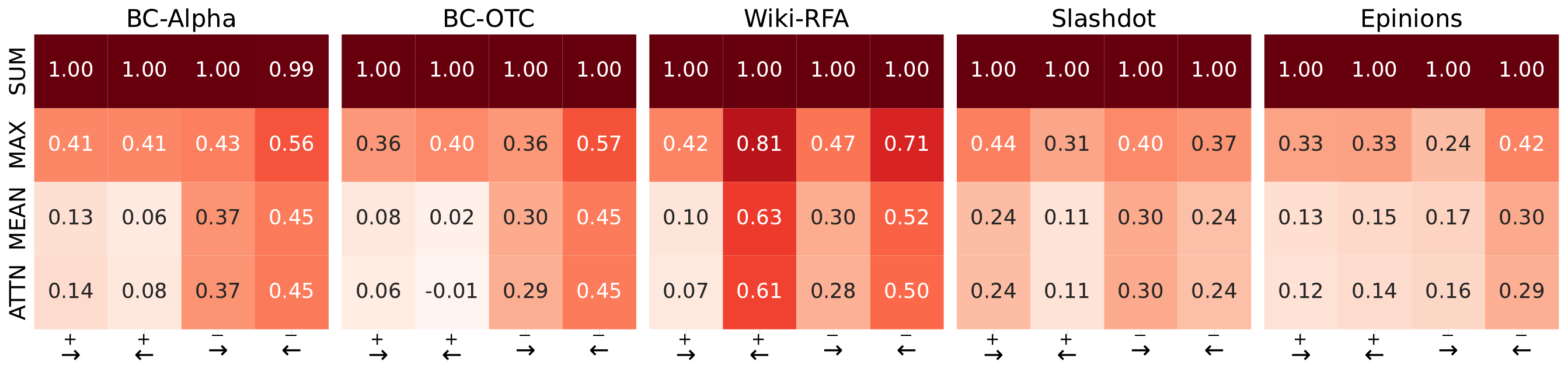}
    \caption{
    Correlations between the norm of the concatenated message of each aggregator (sum, max, mean, and attention) and the degree of $\delta$-neighbors where $\delta \in \mathcal{D} = \set{\rpa, \rna, \lpa, \lna}$.
    }
    \label{fig:experiments:degree}
\end{figure}
\subsection{Further Inspection of Aggregators}
\label{appendix:degree}
We inspect the ability of aggregators such as sum, mean, max, and attention to effectively capture information related to node degrees.
For each $\delta \in \mathcal{D}$ and node $u \in \mathcal{V}$, we measure the correlation between the norm of $\bigparallel_{k \in K}\msg{u}{k}{\delta}$ and the degree $\card{\neighD_{u}}$ where $\msg{u}{k}{\delta}$ is the message obtained from an aggregator. 
The results are demonstrated in Figure~\ref{fig:experiments:degree} for all datasets where as the correlation approaches $+1$, the norm of the concatenated message becomes distinguishable based on the node degree.
Note that only the sum aggregator demonstrates a nearly perfect positive correlation. 
This discrepancy arises because the normalization techniques used by the other aggregators introduce a smoothing effect on the local neighboring structure. 
In contrast, the sum aggregator avoids this smoothing effect, enabling it to capture the local structure more effectively.
Thus, the sum aggregator effectively distributes the degree information throughout the messages, resulting in more expressive features compared to the others when learning the signed graphs.


\subsection{Effect of Number of Factors}
\label{appendix:eff_fac}
\begin{figure}[!ht]
    \centering
    \hspace{5mm}\includegraphics[width=0.8\linewidth]{SENSITIVITY_LEGEND.pdf}\\
    \subfigure[$d_{out} = 64$]{
        \includegraphics[width=0.240\linewidth]{SENSITIVITY_NUM_FACTORS}
        \label{fig:experiments:sensitivity:fac64}
    }
    \hspace{3mm}
    \subfigure[$d_{out} = 128$]{
        \includegraphics[width=0.240\linewidth]{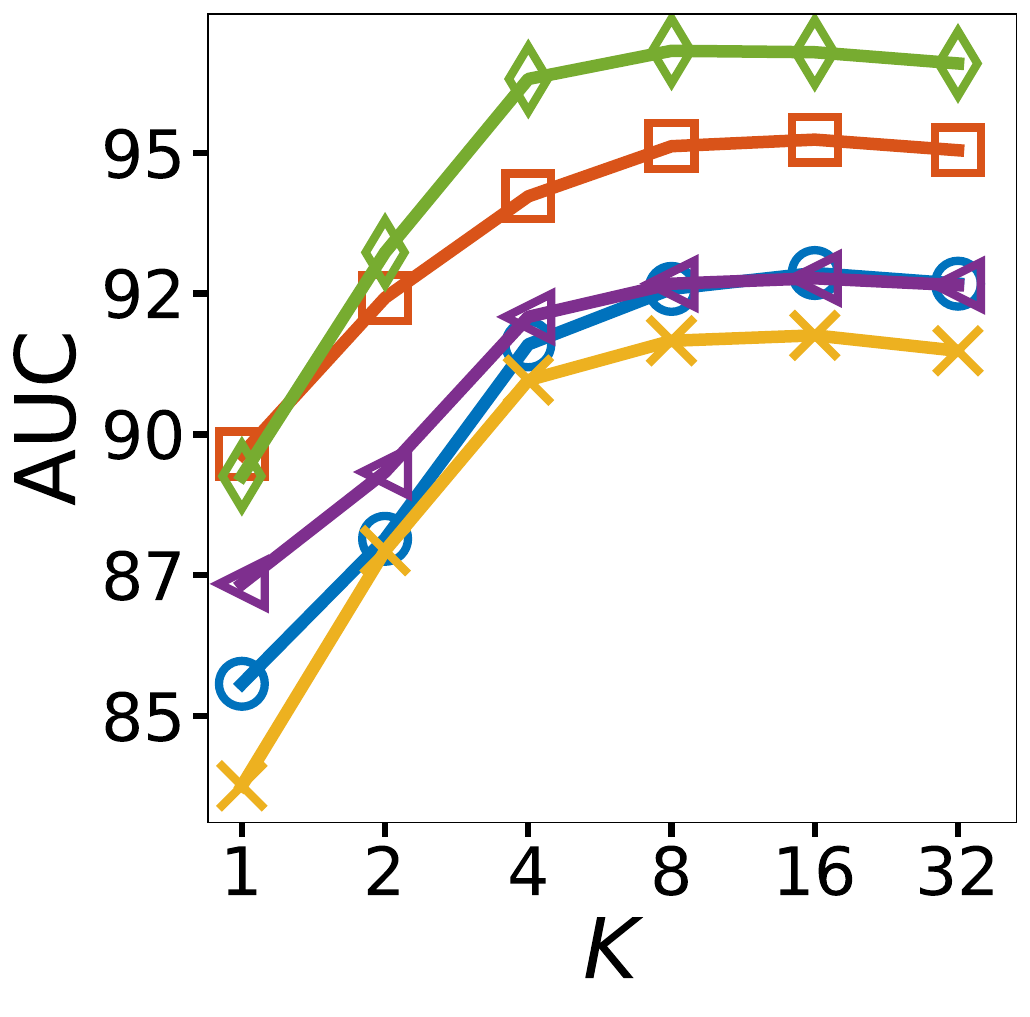}
        \label{fig:experiments:sensitivity:fac128}
    }
    \hspace{3mm}
    \subfigure[$d_{out} = 256$]{
        \includegraphics[width=0.240\linewidth]{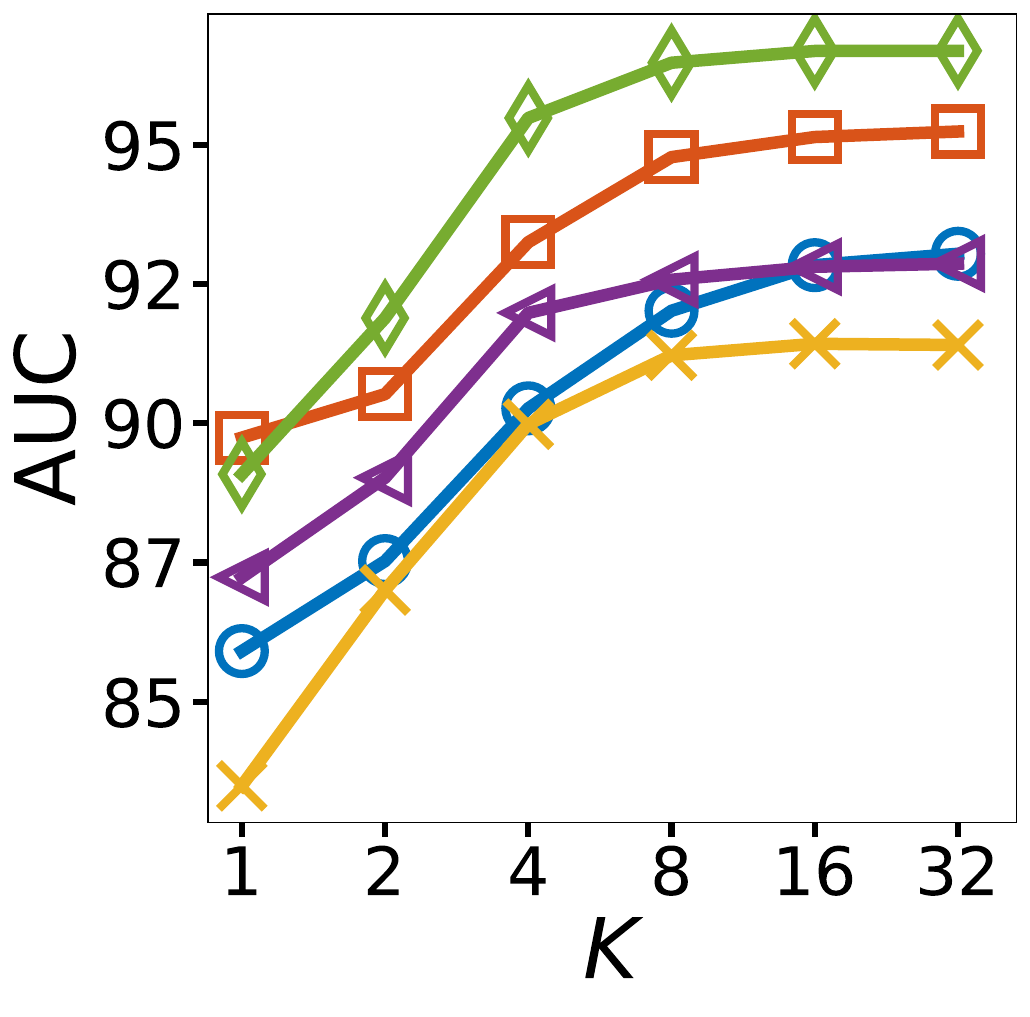}
        \label{fig:experiments:sensitivity:fac256}
    }
    \caption{\vspace{-2mm}
        \label{fig:experiments:sensitivity:fac_dim}
        Effects of the number $K$ of factors  of \method in the different dimension $d_{out}$.
    }
   
\end{figure}

We investigate the effect of the number $K$ of factors in the different dimension $d_{out}$.
When $d_{out}=64$, the performance of AUC initially improves as the value of $K$ increases while the accuracy starts to decrease after $K=16$, as shown in Figure~\ref{fig:experiments:sensitivity:fac64}.
The reasons for this are twofold: firstly, $\frac{d_{out}}{K}$ is too small to adequately model the features of each factor, and secondly, it becomes challenging to properly capture the information from neighboring nodes within the limited feature size.
However, by increasing the dimension $d_{out}$ and enlarging the size of each factor's feature, the performance consistently improves as $K$ increases, as shown in Figures~\ref{fig:experiments:sensitivity:fac128}~and~\ref{fig:experiments:sensitivity:fac256}.

\end{document}